\documentclass{article}

\usepackage[preprint]{customstyle}

\usepackage{multirow}
\usepackage{makecell}
\usepackage{booktabs}
\usepackage{tabularx}
\usepackage{rotating}
\usepackage{subcaption}
\usepackage{wrapfig}
\usepackage{longtable}
\usepackage{xurl}

\usepackage{authblk}
\makeatletter
\renewcommand\AB@affilsepx{\protect\quad\protect\Affilfont}
\makeatother
\renewcommand{\Affilfont}{\small}

\usepackage{pifont}

\usepackage{xcolor}

\usepackage{amsmath,amsfonts,bm}

\def\eqref#1{equation~\ref{#1}}

\def\1{\bm{1}}

\DeclareMathAlphabet{\mathsfit}{\encodingdefault}{\sfdefault}{m}{sl}
\SetMathAlphabet{\mathsfit}{bold}{\encodingdefault}{\sfdefault}{bx}{n}

\usepackage{hyperref}
\usepackage{cleveref}

\definecolor{NewBlueDark}{RGB}{0,45,114}
\definecolor{NewBlue}{RGB}{0,114,188}

\providecommand{\runningtitle}[1]{}

\hypersetup{
  colorlinks=true,
  citecolor=NewBlueDark,
  linkcolor=NewBlueDark,
  urlcolor=NewBlue
}

\newcommand{\taskicon}[1]{%
  \raisebox{-0.3em}{\includegraphics[height=1.4em]{#1}}%
  \hspace{0.1em}%
}

\newcolumntype{P}[1]{>{\raggedright\arraybackslash}p{#1}}

\newcommand{\tool}[1]{\nolinkurl{#1}}

\renewcommand{\arraystretch}{1.18}

\newcommand{\best}[1]{\textbf{\underline{#1}}}

\newenvironment{tooltable}[2]{%
  \small
  \begin{longtable}{@{}P{.32\textwidth}P{.38\textwidth}P{.24\textwidth}@{}}
  \caption{#1}\label{#2}\\
  \toprule
  \textbf{Tool / Action (Task)} &
  \textbf{Tool Description} &
  \textbf{Device / Information Source} \\
  \midrule
  \endfirsthead
  \multicolumn{3}{l}{\textit{\tablename\ \thetable\ continued}}\\
  \toprule
  \textbf{Tool / Action (Task)} &
  \textbf{Tool Description} &
  \textbf{Device / Information Source} \\
  \midrule
  \endhead
  \midrule
  \multicolumn{3}{r}{\textit{Continued on next page}}\\
  \endfoot
  \bottomrule
  \endlastfoot
}{%
  \end{longtable}%
}

\title{Words to World: Benchmarking LLMs in Interactive Physical Environments}

\author[1]{\mbox{Xinnan Dai$^{*}$}}
\author[1]{\mbox{Kaiqi Yang$^{*}$}}
\author[1]{\mbox{Yaxin Li$^{*}$}}
\author[2]{\mbox{Cheng Luo}}
\author[3]{\mbox{Wanli Li}}
\author[4]{\mbox{Mingqing Xiao}}
\author[4]{\mbox{Xinyang Jiang}}
\author[5]{\mbox{Daoan Zhang}}
\author[2]{\mbox{Kaitao Song}}
\author[2]{\mbox{Fengze Liu}}
\author[6]{\mbox{Zefan Cai}}
\author[4]{\mbox{Caihua Shan$^{\dagger}$}}

\affil[1]{Michigan State University}
\affil[2]{Independent Researcher}
\affil[3]{Zhejiang University}

\makeatletter
\renewcommand\AB@affilsep{\protect\\\protect\Affilfont}
\makeatother

\affil[4]{Microsoft}
\affil[5]{University of Rochester}
\affil[6]{University of Wisconsin--Madison}

\begin{document}

\maketitle
\begingroup
\renewcommand{\thefootnote}{\fnsymbol{footnote}}
\footnotetext[1]{Equal contribution.}
\footnotetext[2]{Corresponding author. Email: \texttt{caihuashan@microsoft.com}}
\endgroup

\begin{abstract}
Words are increasingly becoming an interface for action: large language models (LLMs) can now invoke tools, operate devices, and make decisions that affect the physical world. However, current benchmarks largely evaluate LLMs through static text responses, offline tool calls, or simulated tasks, leaving open the question of how well they can act in real physical environments. We present a benchmark for measuring LLMs in closed-loop physical interaction, where models control a harness system connected to real-world components including robotic arms, cameras, and sensors. In each task, the model must issue instructions, receive physical feedback from the environment, interpret the resulting state, and adapt its next action accordingly. This setting evaluates capabilities beyond language understanding, including grounded perception, action planning, tool selection, spatial reasoning, feedback-driven correction, and robustness to execution uncertainty. Through experiments with representative LLMs, we characterize their ability to translate linguistic reasoning into physical actions and identify common failure modes such as hallucinated state changes, poor recovery from unexpected outcomes, and unstable long-horizon plans. Our benchmark aims to provide a systematic foundation for evaluating LLMs as interactive agents that connect words, tools, and the physical world.
\end{abstract}

\section{Introduction}

Large language models (LLMs) increasingly accomplish complex, long-horizon tasks through agent harnesses. Recent achievements span coding and mathematical research: a Claude agent built a Rust-based C compiler capable of compiling the Linux kernel~\cite{carlini2026compiler}, while OpenAI reported a solution to the Navier--Stokes Millennium Prize Problem~\cite{openai2026navierstokes}.
These systems rely not only on model capabilities but also on the harness, which turns model-generated proposals into an iterative process of execution, evaluation, and revision. It coordinates tool execution, preserves intermediate artifacts and task context, and feeds errors and results back into subsequent decisions~\cite{react2023yao,sweagent2024yang}.

\begin{figure}[t]
    \centering
    \makebox[\linewidth][c]{%
    \includegraphics[width=1.15\linewidth]{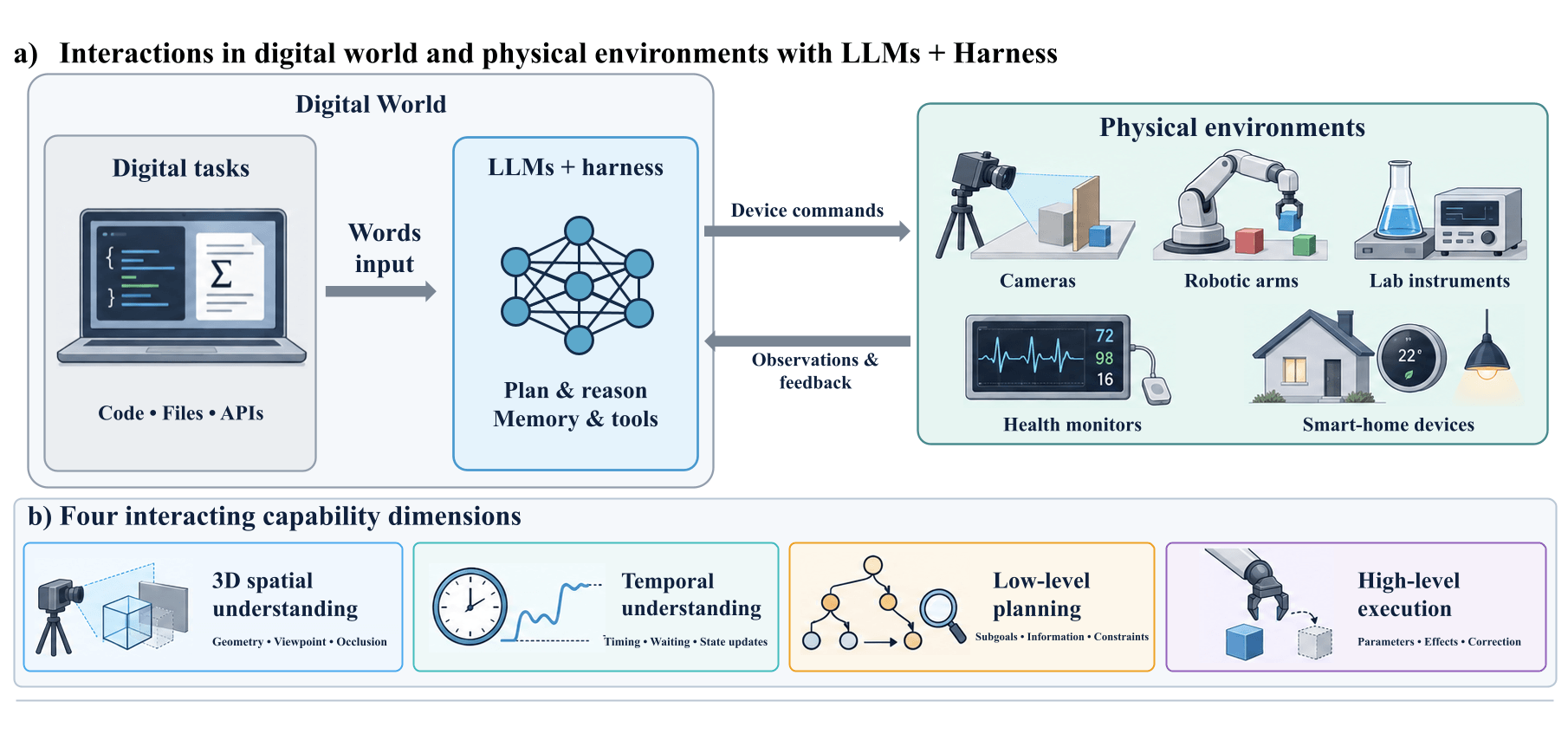}
    }
    
    \captionsetup{
        width=\linewidth,
        justification=justified,
        singlelinecheck=false
    }
    \caption{\textbf{Words to World (W2W-Bench): from digital task completion to physical interaction.}
    Tool access alone does not ensure reliable task completion.
    a) harness systems must close the loop between device commands,
    observations, and verification of intended effects.
    b) W2W-Bench assesses this loop across four interacting
    dimensions---3D spatial understanding, temporal understanding,
    high-level planning, and low-level execution.}
    \label{fig:w2w-overview}
\end{figure}

These successes in digital environments raise a natural question: \emph{how reliably can current model--harness systems complete language-specified tasks in physical environments?}
Figure~\ref{fig:w2w-overview}(a) illustrates the interaction setting. Task instructions enter a model--harness system operating in the digital world, where reasoning, memory, and tool orchestration support its decisions. Device interfaces connect these decisions to physical environments: the system issues commands and receives observations and feedback that inform subsequent reasoning and action.
Exposing a device through a tool interface makes its operations accessible, but does not ensure that the system can acquire informative observations, interpret the resulting feedback, or verify the intended effects. Reliable task completion therefore depends on maintaining this interaction loop, using feedback to update state estimates, revise plans, and correct execution.

Recent benchmarks investigate parts of this problem: EmbodiedBench~\cite{embodiedbench2025yang} and RoboCerebra~\cite{robocerebra2025han} evaluate embodied planning, perception, and execution, while SimuHome~\cite{simuhome2026seo} and DevicesWorld~\cite{devicesworld2026li} examine dynamic device operation and cross-device coordination. Building on these efforts, we apply a shared diagnostic framework to investigate whether---and under what interaction conditions---the same capability bottlenecks recur across diverse tasks and environments.

We introduce \textbf{Words to World (W2W-Bench)}, a benchmark comprising \textbf{11 task families} for systematically assessing complete model--harness systems on language-specified tasks involving physical environments.
As shown in Figure~\ref{fig:w2w-overview}(b), we organize these tasks around four complementary capability dimensions that provide a shared basis for diagnosing behavior throughout the interaction loop.

\textbf{3D spatial understanding} concerns the interpretation of geometry, viewpoint, occlusion, and relative pose from incomplete observations.
\textbf{Temporal understanding} concerns evolving signals and states, including when to observe, how long to wait, and when an earlier observation must be updated.
\textbf{Low-level execution} concerns translating selected subgoals into device commands and parameters, inspecting their effects, and correcting deviations.
\textbf{High-level planning} concerns the decomposition of objectives into subgoals by models themslves, the selection of informative observations, and the coordination of actions under task and resource constraints.

The task families span visual capture and generation, active scanning and recognition, experimental execution and construction, document and time-series analysis, and dynamic device control.
Representative device interfaces provide access to controllable cameras, robotic arms, laboratory instruments, health-monitoring tools, and smart-home devices.
Across these settings, we assess both \emph{task-specific outcomes} and \emph{process-level diagnostics}: the former measure whether task requirements are satisfied, while the latter characterize the intermediate behavior connecting observation, planning, execution, and verification. This combination supports analysis of both task completion and recurring difficulties within the interaction loop.

Our evaluation shows that strengths vary by model, harness, and task, revealing distinct performance profiles. GPT-6 Astra can excel at planning and execution while struggling with basic 3D spatial or temporal understanding. Opus 5 with Claude Code does not dominate any single task but performs best overall. We also systematically analyze failure cases to examine the gap between language-based reasoning and real-world execution, focusing on how spatial and temporal errors interact to produce broader failure patterns. Our contributions are threefold:
\begin{enumerate}
    \item We formulate a capability-oriented framework for evaluating model--harness systems in physical interaction through four complementary dimensions, organizing heterogeneous tasks by their primary demands while accounting for dependencies between capabilities.
    \item We present W2W-Bench, a suite of 11 task families spanning heterogeneous observations and device interfaces, with task-specific outcome metrics and process-level diagnostics.
    \item We evaluate multiple model--harness configurations and identify complementary strengths, recurring failure patterns, and task-dependent differences between harnesses.
\end{enumerate}

\section{W2W-Bench}

\begin{figure*}[t]
  \centering

  \makebox[\linewidth][c]{%
    \includegraphics[width=1.2\linewidth]{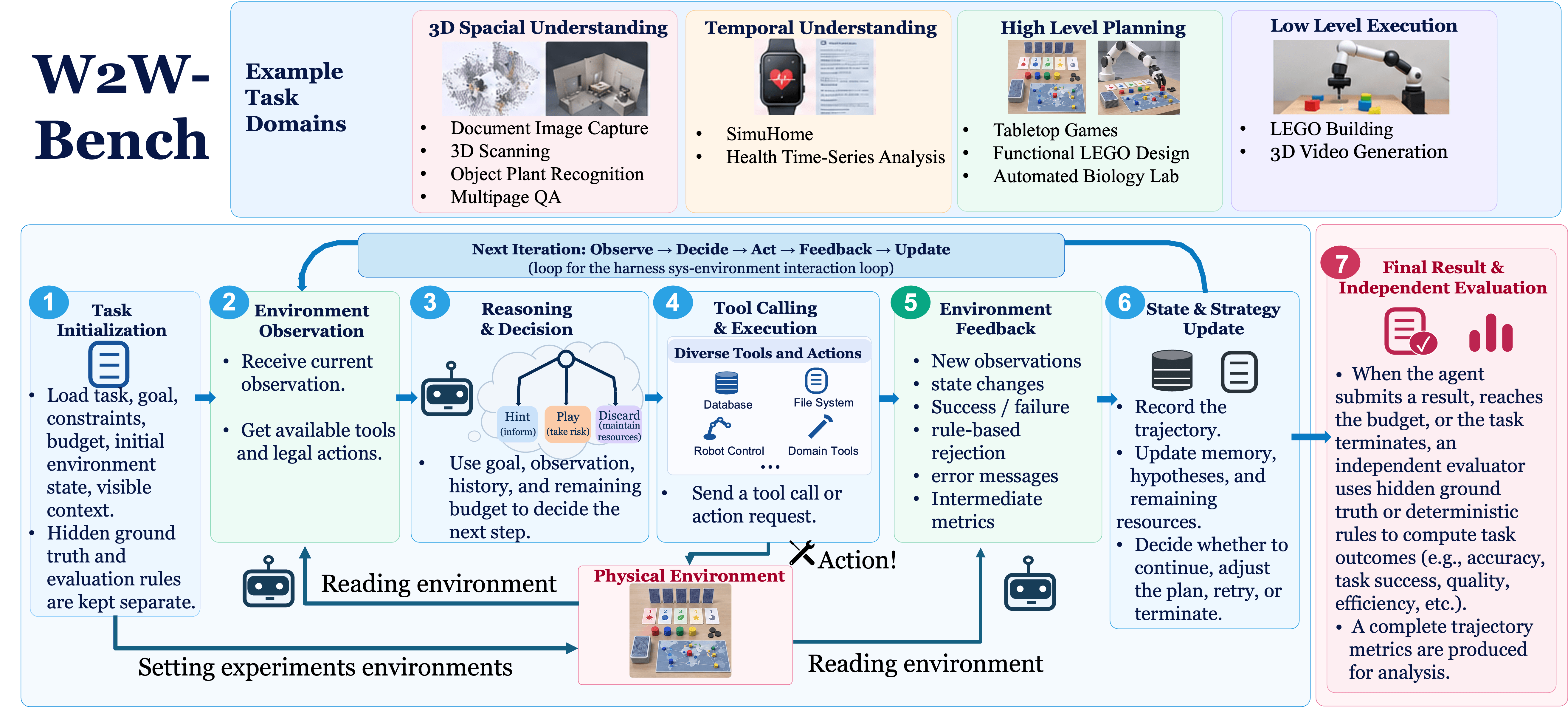}%
  }

  \captionsetup{
    width=\linewidth,
    justification=justified,
    singlelinecheck=false
  }
  \caption{Overview of W2W-Bench. We evaluate how LLMs interact with the physical world.}
  \label{fig:harness_benchmark_overview}
\end{figure*}

W2W-Bench evaluates LLM--harness systems across four capability domains: 3D spatial understanding, temporal understanding, high-level planning, and low-level execution. Despite differences in tools and environments, all tasks follow an interaction loop, as shown in Figure~\ref{fig:harness_benchmark_overview}.

At each step, the LLM uses the task goal, current observations, and interaction history to decide what information to acquire or what action to take. The harness executes the corresponding tool call in the environment, which returns new observations, state changes, or execution errors. The system uses this feedback to update its understanding and choose the next step. The benchmark records the complete trajectory of this process.

An independent evaluator remains outside the interaction loop: hidden ground truth and scoring rules are not available to the LLM--harness system. After the episode, it assesses both \emph{task-specific outcomes} and \emph{process-level diagnostics}. The latter characterize how the system acquired evidence, executed actions, responded to feedback and errors, and used its interaction budget.

Specifically, we describe the 11 physical interaction tasks in~\Cref{sub_sec:11tasks}, the tools used in~\Cref{subsec:tool_use}, and the datasets, execution environments, and evaluation metrics in~\Cref{subsec:evaluation_setting}.

\subsection{11 Tasks for Physical Interaction}
\label{sub_sec:11tasks}
We organize the benchmark tasks into four categories spanning diverse observation modalities, tool interfaces, action spaces, and environments. Together, they allow us to evaluate how different harnesses support perception, planning, information gathering, and interaction.

\paragraph{3D Spacial Understanding}
3D spatial understanding concerns how a LLMs control the vision device to observe the physical words, such as geometry, viewpoint, occlusion, and relative pose from incomplete visual observations. We evaluate the complete LLM--harness system in each task.
\mbox{\textbf{\taskicon{figure/icons/icon\_doc.png}{Document Image Capture.}}}
The goal is to produce an identity-document image that meets all quality requirements despite occlusions, shadows, and glare. After each capture, the system adjusts the camera settings based on visual feedback and selects the final image. \mbox{\textbf{\taskicon{figure/icons/icon\_3Dscan.png}{3D Scanning.}}}
Accurate reconstruction requires sufficient coverage of the object's surface. Reconstruction feedback guides the choice of new viewpoints to reveal occluded regions without making redundant scans. \mbox{\textbf{\taskicon{figure/icons/icon\_object.png}{Object Recognition.}}}
A partly occluded 3D object must be labeled correctly. The system changes camera azimuth, elevation, and distance to gather complementary views and resolve ambiguity. \mbox{\textbf{\taskicon{figure/icons/icon\_QA.png}{Multipage Question Answering.}}}
The system answers four questions about a 20-page document from MP-DocVQA \citep{tito2023hierarchical}. Because supporting evidence spans one to four pages (2.75 on average), it must choose which pages or regions to inspect and locate evidence among irrelevant content.

\paragraph{Temporal Understanding}
Temporal understanding focuses on information that evolves over time and whose interpretation depends on when and over what duration it is observed. 
The harness therefore provides temporally grounded observations and interfaces through which the model can decide when additional information is needed and which signals or states should be acquired. 
\mbox{\textbf{\taskicon{figure/icons/icon\_health.png}{Health Time-Series Analysis.}}}
The goal is to infer an underlying health condition from incomplete, noisy physiological signals. The system selectively requests measurements or future observation windows, then uses each result to decide what evidence is needed next. The challenge is to identify informative signals without acquiring unnecessary observations.
\mbox{\textbf{\taskicon{figure/icons/icon\_simu.png}{SimuHome.}}}
The goal is to complete household objectives as environmental conditions change over time. In our Matter-style SimuHome adaptation \citep{simuhome2026seo}, the system queries relevant state through a GUI, controls devices through constrained actions, and checks their effects. The challenge is to choose and time actions using only partially observed state.

\paragraph{Low Level Execution}
At the level of low-level execution, the primary challenge shifts from deciding what should be accomplished to determining how a selected action or subgoal should be physically realized. The harness grounds model decisions in concrete device
commands and control parameters while exposing the resulting effects in the environment.

\mbox{\textbf{\taskicon{figure/icons/icon\_3Dvideo.png}{3D Video Generation.}}}
The goal is to produce a high-quality 360-degree product video with informative, complementary views. The system inspects rendered views and adjusts camera radius, height, elevation, and focus before selecting the final video. The challenge is to maintain visual quality and viewpoint diversity across the sequence.
\mbox{\textbf{\taskicon{figure/icons/icon\_lego1.png}{LEGO Building.}}}
The goal is to assemble a specified, correctly connected and stable LEGO structure in a simulated environment. The system places pieces, inspects the result through a wrist camera, and corrects errors before continuing. Small pose errors, failed connections, or collisions can obstruct later steps.

\paragraph{High Level Planning}
High-level planning addresses the challenge of turning a task objective into a sequence of intermediate decisions under evolving environmental and task constraints. 
The model must use this information to decompose objectives, select appropriate subgoals or strategies, and revise earlier decisions when execution reveals new constraints or unexpected outcomes.

\mbox{\textbf{\taskicon{figure/icons/icon\_lab.png}{Automated Biology Laboratory.}}}
The goal is to complete an experiment through operations such as pipetting, dilution, and measurement. The system uses visual and GUI feedback to check each result and adjust subsequent operations. The challenge is to maintain correct liquid volumes, device states, and action order, since an early error can affect later steps. \mbox{\textbf{\taskicon{figure/icons/icon\_game.png}{Tabletop Games.}}}
In Hanabi, Regicide, and Pandemic, the goal is to achieve each game's objective as its state evolves. The system executes actions through a MuJoCo robot and replans from the resulting game and physical states. Locally valid moves may have poor long-term consequences. \mbox{\textbf{\taskicon{figure/icons/icon\_lego2.png}{Functional LEGO Design.}}}
Given a functional objective and a library of digital LEGO components, the goal is to design a structure that works and satisfies geometric constraints. The system selects components, inspects design feedback, and revises their arrangement. The challenge is to ensure stability, connectivity, and clearance while choosing among multiple possible designs.

\subsection{Tool use in physical world}
\label{subsec:tool_use}

\begin{wrapfigure}{r}{0.5\linewidth}
  \centering
  \includegraphics[width=\linewidth]{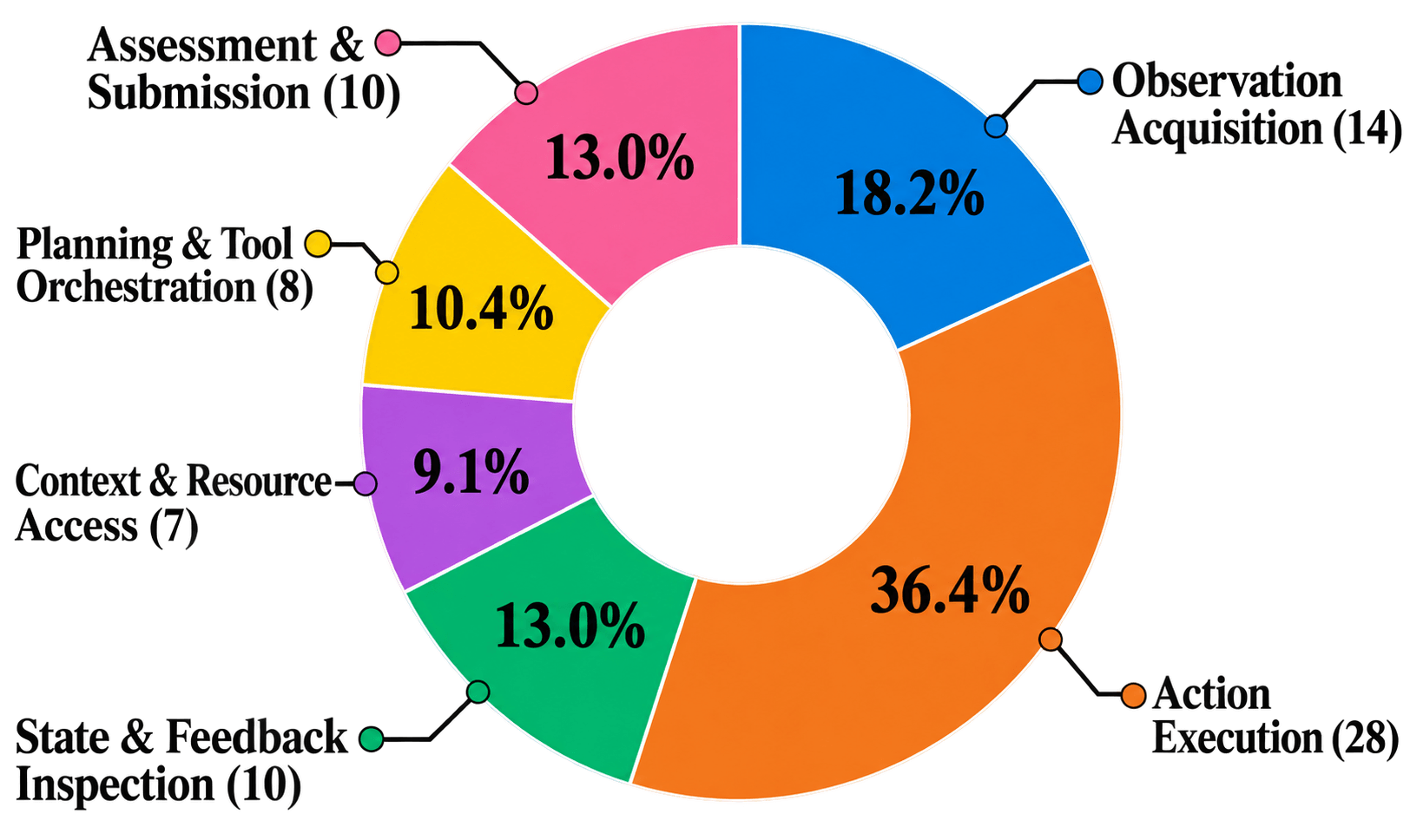}
  \caption{Tool Usage by Category}
  \label{fig:tool_distribution}
\end{wrapfigure}

To interact with physical or simulated environments, LLMs use harness-provided tools to observe, act, and track changes over time. We group these tools by their primary function, as shown in~\Cref{fig:tool_distribution}. \textbf{Observation Acquisition} interfaces collect task evidence, such as camera views, document regions, physiological signals, and laboratory measurements. \textbf{Context \& Resource Access} interfaces provide supporting information, including clinical records, coding dictionaries, reconstruction targets, LEGO manuals, and environmental rules. \textbf{Planning \& Tool Orchestration} interfaces help initialize tasks, choose observation windows, validate action parameters, and schedule workflows. \textbf{Action Execution} interfaces then carry out the selected operations, such as device commands, LEGO manipulation, laboratory procedures, and legal Tabletop game moves. After execution, \textbf{State \& Feedback Inspection} interfaces reveal current states, intermediate results, and progress. The model can use this feedback to verify outcomes, revise its plan, or recover from errors. Finally, \textbf{Assessment \& Submission} interfaces record interpretations and evidence, submit answers or artifacts, and terminate episodes when required.

\subsection{Evaluation Setting}
\label{subsec:evaluation_setting}

\paragraph{Dataset}The dataset draws on several open-source resources, including 3D models from ModelNet10~\cite{wu20153dshapenetsdeeprepresentation} and ABO20~\cite{collins2022abodatasetbenchmarksrealworld}, documents from DocVQA~\cite{tito2023hierarchicalmultimodaltransformersmultipage}, and health data from MIMIC-III~\cite{PhysioNet-mimiciii-demo-1.4, pollard_physionet_2026} and AutoBio~\cite{lan2026autobio}. We also use publicly available LEGO parts~\footnote{https://library.ldraw.org/parts/list} and filming guidelines~\footnote{https://jimmccarthy.org/wp-content/uploads/2016/06/checklist-fiveshot-0.3-small.pdf}. Using these resources, we randomly generate 10--30 evaluation cases for each LLM–harness system. Details are provided in Appendix~\Cref{app_sec:dataset}.

\paragraph{Running Environments}

Experiments run on Ubuntu 24.04 with 24 AMD EPYC CPU cores, 216~GiB RAM, and one NVIDIA A100 GPU (80~GiB). Visual physics tasks use MuJoCo 3.x with GPU rendering; 3D reconstruction uses MuJoCo 3.11 and BlenderProc 2.8 for procedural scene generation. In evaluation, we test harness include Codex-0.149 for gpt-5.6-sol/luna/terra Codex-0.153 for gpt-6-astra, Claude Code2.1.173 for Opus5, Pi 0.84.4~\footnote{https://github.com/earendil-works/pi/releases}, Hermes 0.21.0~\footnote{https://hermes-agent.nousresearch.com/docs/}, and Factory Dorid 0.208.2~\footnote{https://factory.com/}.

\paragraph{Metrics}

Evaluation combines task-specific outcomes with process-level diagnostics. Task-specific outcomes assess the final result when a task has a defined target. For example, 3D scanning uses the F-score common in point-cloud evaluation to measure how closely a generated 3D object matches the target. Tabletop Games reports a separate score for each of its three games rather than averaging across games. Within each task or game, we report the mean score across episodes.

Process-level diagnostics assess whether individual steps execute successfully. Depending on the task, the metric is either a binary success indicator or a continuous score. Let \(N\) be the number of episodes, \(b_i\in\{0,1\}\) the strict success indicator for episode \(i\), and \(s_i\in[0,1]\) a task-specific continuous score. We aggregate these metrics as
\begin{equation}
    \operatorname{PassRate}=\frac{1}{N}\sum_{i=1}^{N} b_i,
    \qquad
    \operatorname{MeanScore}=\frac{1}{N}\sum_{i=1}^{N} s_i.
    \label{eq:common_aggregation}
\end{equation}
The full details about evaluations are shown in Appendix~\Cref{{app:metric}}.

\section{Experiments}
\subsection{The Evaluation of Harness in Physical interaction}

\textbf{Main Results}

Table~\ref{tab:main-harness-llm-results} presents the main results. Across the 11 tasks, gpt-6-astra and claude-opus-5 occupy the top two positions overall, but their strengths differ. Claude-opus-5 performs consistently well across nearly all tasks, whereas gpt-6-astra is particularly strong in planning and execution but weaker in basic sensing such as 3D spatial and temporal understanding. This pattern suggests that gpt-6-astra’s overall ranking is driven largely by its performance on action-oriented tasks.

\begin{table*}[t]
\centering

\captionsetup{
  width=0.95\textwidth,
  justification=raggedright,
  singlelinecheck=false
}
\caption{Main-experiment results across harness systems and LLMs. All values are percentages. Bold and underlined values indicate the best and second-best results for each task, respectively.}
\label{tab:main-harness-llm-results}

\footnotesize
\setlength{\tabcolsep}{3.2pt}
\renewcommand{\arraystretch}{1.12}

\resizebox{0.95\textwidth}{!}{%
\begin{tabular}{@{}llcccccccc@{}}
\toprule
\multirow{2}{*}{\textbf{Harness}} &
\multirow{2}{*}{\textbf{LLM}} &
\multicolumn{5}{c}{\textbf{3D Spatial Understanding}} &
\multicolumn{3}{c}{\textbf{Temporal Understanding}} \\
\cmidrule(lr){3-7}
\cmidrule(lr){8-10}
& &
\makecell{Document\\Capture} &
\makecell{3D\\Scanning} &
\makecell{Object\\Recognition} &
\makecell{Multipage\\QA} &
\makecell{Avg.\\rank} &
\makecell{Health\\Time-Series} &
\makecell{SimuHome} &
\makecell{Avg.\\rank} \\
\midrule
\multirow{4}{*}{Codex CLI}
& gpt-5.6-sol
& \best{95.00} & 60.45 & 40.00 & 64.38 & 9.25
& 53.33 & \underline{95.15} & 5.50 \\
& gpt-5.6-luna
& \best{95.00} & 62.94 & 45.00 & 52.81 & 8.00
& 50.00 & 89.06 & 9.00 \\
& gpt-5.6-terra
& \best{95.00} & 62.49 & 60.00 & 60.31 & 6.50
& 56.67 & 91.92 & 6.50 \\
& gpt-6-astra
& {90.00} & 62.68 & \best{80.00} & 67.19 & 5.75
& 44.83 & 92.40 & 8.50 \\
\midrule
Claude Code
& claude-opus-5
& \best{95.00} & 63.11 & 55.00 & \underline{72.19} & \best{3.75}
& \underline{63.33} & 93.61 & \best{2.75} \\
\midrule
\multirow{3}{*}{Pi}
& claude-opus-5
& \best{95.00} & 63.21 & 55.00 & 70.72 & \best{3.75}
& 55.56 & 91.15 & 7.50 \\
& gpt-5.6-sol
& \best{95.00} & 62.26 & 50.00 & 65.63 & 7.00
& 58.62 & 92.81 & {5.00} \\
& gpt-6-astra
& 85.00 & \best{77.91} & \underline{75.00} & 70.00 & {4.62}
& 60.00 & \best{99.05} & \best{2.75} \\
\midrule
\multirow{2}{*}{Hermes}
& claude-opus-5
& \best{95.00} & \underline{63.44} & 50.00 & 64.69 & 5.62
& \underline{63.33} & 75.43 & 5.75 \\
& gpt-5.6-sol
& {90.00} & 61.91 & 35.00 & 64.69 & 10.12
& 48.28 & 67.30 & 10.50 \\
\midrule
\multirow{2}{*}{Factory Droid}
& claude-opus-5
& {90.00} & 63.02 & 50.00 & 69.23 & 6.75
& \best{73.33} & 45.08 & 6.50 \\
& gpt-5.6-sol
& 85.00 & 62.07 & 55.00 & \best{72.81} & 6.88
& 60.00 & 60.51 & 7.75 \\
\bottomrule
\end{tabular}}

\vspace{0.9em}

\resizebox{0.85\textwidth}{!}{%
\begin{tabular}{@{}llcccccccc@{}}
\toprule
\multirow{2}{*}{\textbf{Harness}} &
\multirow{2}{*}{\textbf{LLM}} &
\multicolumn{3}{c}{\textbf{Low-Level Execution}} &
\multicolumn{4}{c}{\textbf{High-Level Planning}} &
\multirow{2}{*}{\makecell{\textbf{Overall}\\\textbf{Avg. rank}}} \\
\cmidrule(lr){3-5}
\cmidrule(lr){6-9}
& &
\makecell{3D\\Video} &
\makecell{LEGO\\Building} &
\makecell{Avg.\\rank} &
\makecell{AutoBio} &
\makecell{Functional\\LEGO} &
\makecell{Tabletop\\H/R/P} &
\makecell{Avg.\\rank} & \\
\midrule
\multirow{4}{*}{Codex CLI}
& gpt-5.6-sol
& \underline{60.00} & 10.00 & 4.25
& 76.45 & 40.00 & \underline{45.00} & 5.17 & 6.55 \\
& gpt-5.6-luna
& 15.00 & 0.00 & 10.25
& 71.43 & 10.00 & 34.00 & 9.33 & 8.95 \\
& gpt-5.6-terra
& 30.00 & 0.00 & 9.50
& 70.94 & 0.00 & 36.78 & 9.33 & 7.82 \\
& gpt-6-astra
& 55.00 & \best{50.00} & \best{3.00}
& 73.80 & \best{80.00} & \underline{45.00} & \underline{3.50} & 5.14 \\
\midrule
Claude Code
& claude-opus-5
& 30.00 & \underline{25.00} & 5.00
& 72.30 & 50.00 & \best{46.33} & 4.67 & \best{4.05} \\
\midrule
\multirow{3}{*}{Pi}
& claude-opus-5
& 10.00 & 0.00 & 10.75
& \underline{82.18} & {70.00} & 36.67 & 4.83 & 6.00 \\
& gpt-5.6-sol
& \underline{60.00} & 12.50 & \best{3.00}
& 81.43 & 40.00 & 20.56 & 7.33 & 6.00 \\
& gpt-6-astra
& \best{70.00} & 10.00 & {3.25}
& 58.97 & \best{80.00} & 21.89 & 8.00 & \underline{4.95} \\
\midrule
\multirow{2}{*}{Hermes}
& claude-opus-5
& 30.00 & 0.00 & 9.50
& 77.86 & 40.00 & 39.22 & 5.67 & 6.36 \\
& gpt-5.6-sol
& \underline{60.00} & 0.00 & 6.25
& \best{85.21} & \best{80.00} & 42.22 & \best{2.33} & 7.36 \\
\midrule
\multirow{2}{*}{Factory Droid}
& claude-opus-5
& 35.00 & 0.00 & 8.25
& 69.43 & {70.00} & 38.00 & 6.83 & 7.00 \\
& gpt-5.6-sol
& 40.00 & 11.11 & 5.00
& 67.96 & 20.00 & 19.11 & 11.00 & 7.82 \\
\bottomrule
\end{tabular}}

\end{table*}

From the perspective of harness systems, the results reveal task-dependent rather than uniform advantages. Selecting the best-performing configuration for each harness, Claude Code achieves the best overall result, followed by Pi and Codex CLI. Across task categories, Claude Code performs best on 3D spatial understanding and ties with Pi on temporal understanding, whereas Codex CLI performs best on low-level execution and high-level planning. Pi ranks second in 3D spatial understanding and low-level execution, while Claude Code ranks second in high-level planning. In contrast, Hermes and Factory Droid perform less strongly across the selected configurations. These results suggest that the harnesses have complementary strengths: Claude Code and Pi are more effective for spatial and temporal understanding, whereas Codex CLI works better for execution and planning.

We further evaluate open-source LLMs, represented by Qwen 3.8 Max, on representative tasks selected from four categories of tasks with the Pi harness.

\begin{wraptable}[10]{r}{0.45\linewidth}
  \centering

  \captionsetup{
    width=\linewidth,
    justification=raggedright,
    singlelinecheck=false
  }
  \caption{Qwen3.8 Max in Physical Environment Interaction. Harness is Pi.}
  \label{tab:opensource-results}

  \scriptsize
  \setlength{\tabcolsep}{5.0pt}
  \renewcommand{\arraystretch}{1.12}
  \resizebox{\linewidth}{!}{%
    \begin{tabular}{lcccc}
      \toprule
      LLM &
      \makecell{Object\\Recognition} &
      \makecell{Simu\\Home} &
      \makecell{LEGO\\Building} &
      \makecell{Functional\\LEGO} \\
      \midrule
      qwen3.8-max
      & \textbf{80.00} & 87.99 & 0.00 & 0.00 \\
      claude-opus-5
      & 55.00 & 91.15 & 0.00 & 70.00 \\
      gpt-5.6-sol
      & 50.00 & 92.81 & \textbf{12.50} & 40.00 \\
      gpt-6-astra
      & 75.00 & \textbf{99.05} & 10.00 & \textbf{80.00} \\
      \bottomrule
    \end{tabular}%
  }
\end{wraptable}

As shown in Table~\ref{tab:opensource-results}, Qwen performs competitively on 3D spatial understanding, suggesting strong visual reasoning capabilities. However, it performs less well on temporal understanding and fails on planning and execution tasks.

\subsection{The gaps for LLMs to adapt in real-world}
We further investigate the bottlenecks that limit LLM-harness-based systems’ ability to interact with the physical world. We begin with sensor use and 3D spatial in~\Cref{subsub_sec:3D_gap} and temporal understanding in~\Cref{subsub_sec:temproal_gap}, then examine how gaps in these foundational capabilities affect more complex strategies and applications for multiple object interaction in physical environment~\Cref{subsub_sec:strategy_impact}.

\subsubsection{3D Spatial Understanding}
\label{subsub_sec:3D_gap}
\begin{figure*}[t]
  \centering

  \begin{minipage}[t]{0.35\textwidth}
    \centering
    \includegraphics[width=\linewidth]{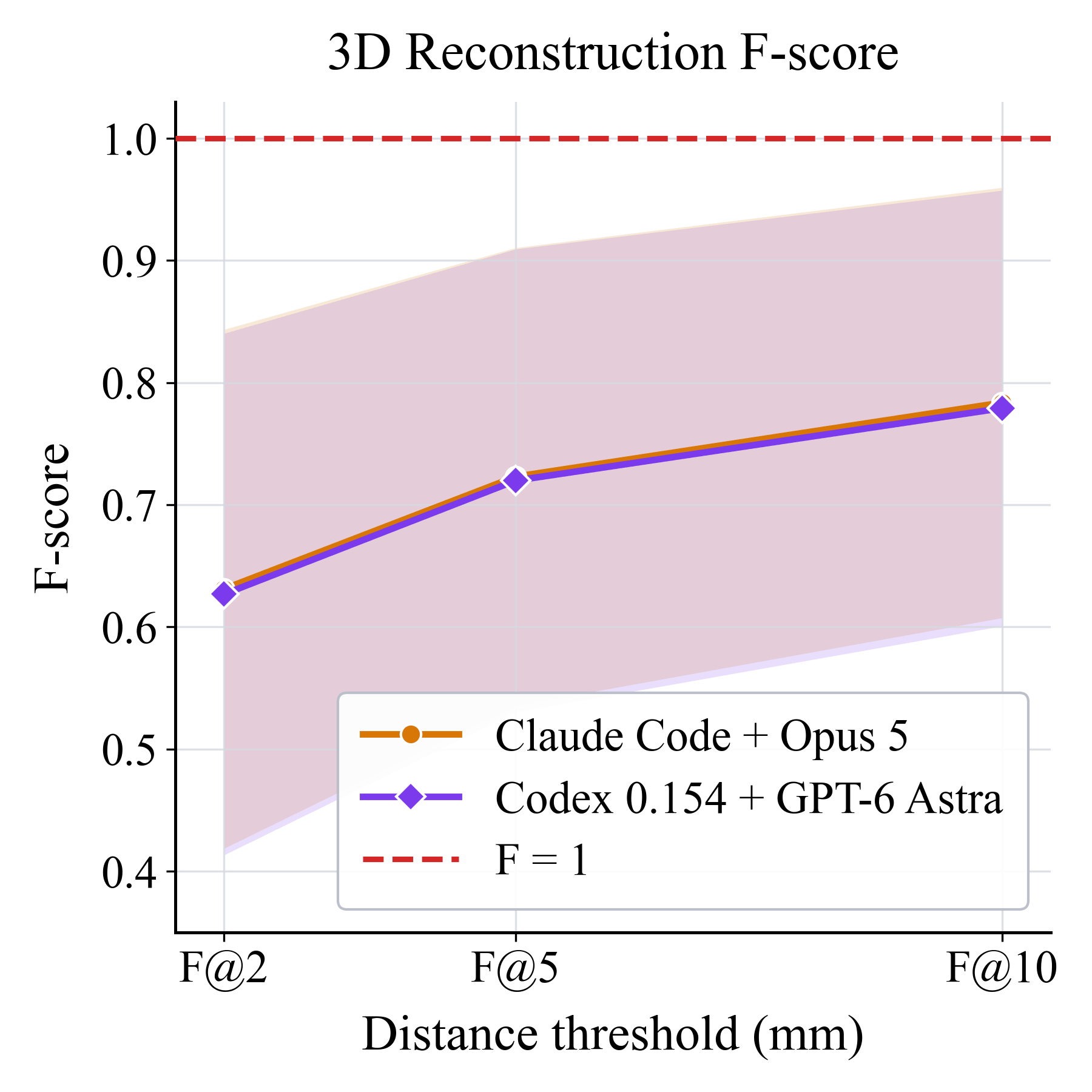}
    \captionsetup{
      width=0.8\linewidth,
      justification=justified,
      singlelinecheck=false
    }
    \captionof{figure}{F-scores at tolerances from 2 to 10 mm.}
    \label{fig:F_score}
  \end{minipage}
  \hfill
  \begin{minipage}[t]{0.3\textwidth}
    \centering
    \includegraphics[width=\linewidth]{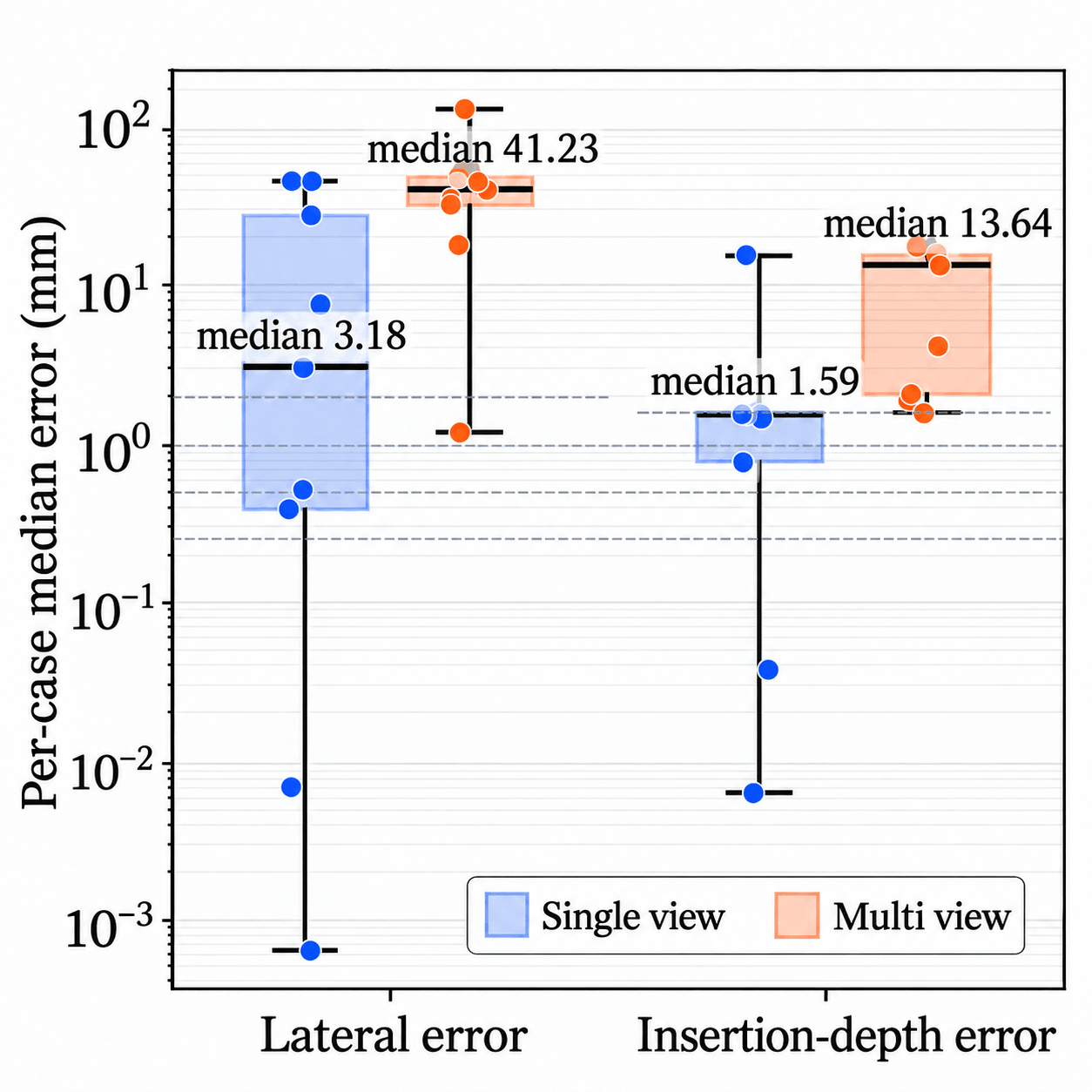}
    \captionsetup{
      width=0.8\linewidth,
      justification=justified,
      singlelinecheck=false
    }
    \captionof{figure}{Error modes under single and multi-views.}
    \label{fig:vis_error}
  \end{minipage}
  \hfill
  \begin{minipage}[t]{0.3\textwidth}
    \centering
    \includegraphics[width=\linewidth]{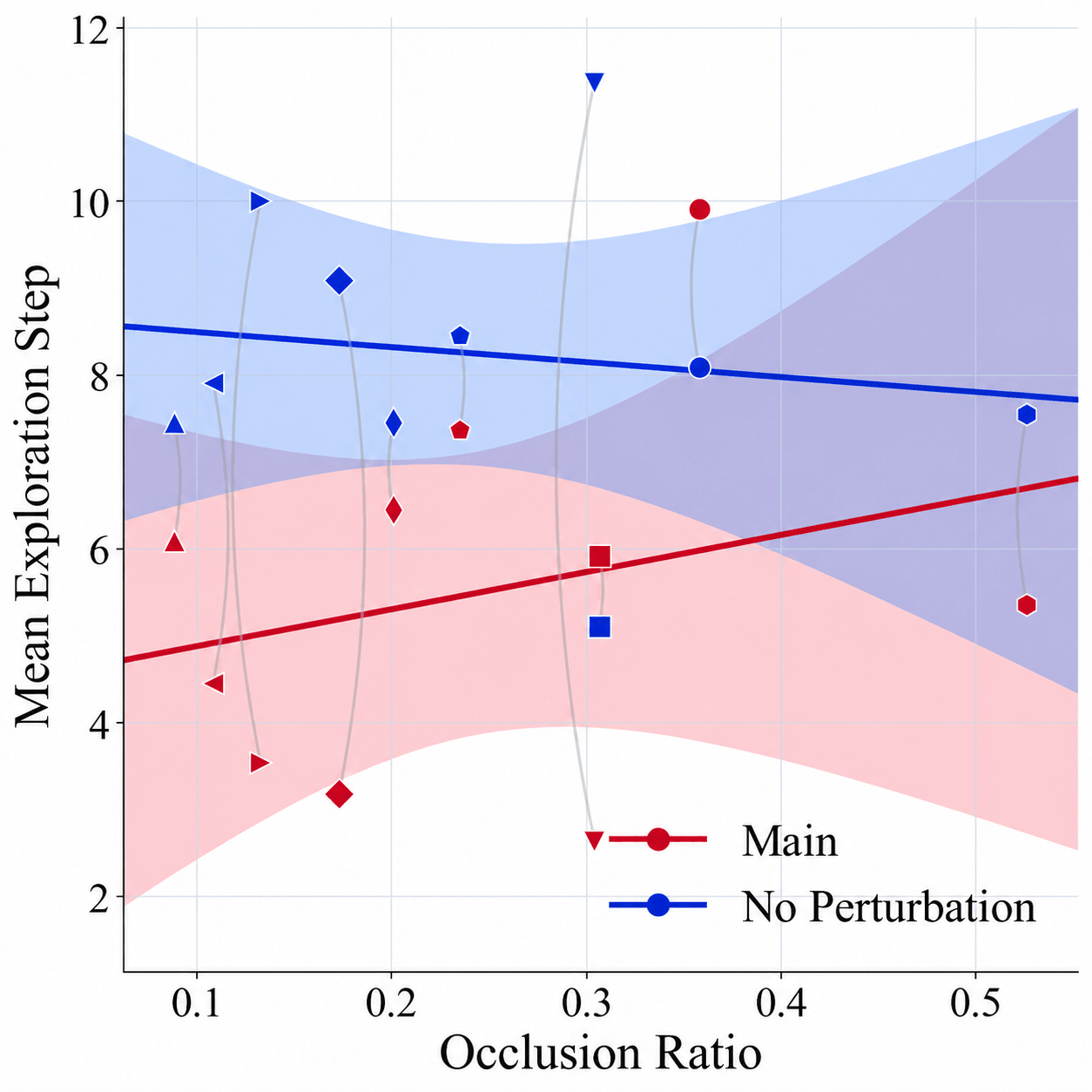}
    \captionsetup{
      width=0.8\linewidth,
      justification=justified,
      singlelinecheck=false
    }
    \captionof{figure}{Mean steps across difficulty levels.}
    \label{fig:difficulty-main}
  \end{minipage}
\end{figure*}

To guide physical actions reliably, LLMs must perceive fine-grained spatial relationships and seek informative views when observations are noisy or incomplete. Failures in these capabilities can leave millimeter-scale misalignments undetected and actions guided by insufficient visual evidence.
\paragraph{Millimeter-scale visual deviations}
Spatial is one of the core capabilities that enables LLMs to interact with the real world. However, our experimental results suggest that LLMs are unable to fully extract information from visual inputs, particularly due to limitations in spatial understanding and robustness to visual perturbations. For example, LLMs still struggle to understand precise spatial relationships. The 3D scanning task achieves a low F-score across different LLMs, even when evaluated with a 10 mm tolerance, as shown in Figure~\ref{fig:F_score}.

These errors directly affect vision-based action tasks. For example, the LEGO building task requires an understanding of the structures and small slot features of individual components. The results suggest that the blocks are merely overlapping rather than properly assembled. Even with multi-view feedback, LLMs are unable to perceive millimeter-level visual deviations as shown in~\Cref{fig:vis_error}.

\paragraph{Visual Information Foraging}

In our experiments, we find that the degree of surrounding perturbation has no significant correlation with exploring steps. Intuitively, when the situation has more perturbation objects and thus the task is more difficult, the exploration should be comprehensive and adequate, regardless of the final prediction. However, as shown in Figure~\ref{fig:difficulty-main}, the perturbation by occlusion ratio (x-axis) exhibits no significant correlation with the number of steps (y-axis), where the blue shaded area denotes the 95\% confidence interval.

This limitation degrades both accuracy and efficiency. In the {object recognition} task, heavier occlusion diminishes single-view information, requiring multi-view exploration across diverse angles to gather sufficient evidence. However, LLMs fail to gauge accumulated evidence or identify stopping thresholds. Consequently, they either under-explore complex cases, leading to task failure, or over-inspect simple ones, causing severe inefficiency.

\subsubsection{Temproal understanding}
\label{subsub_sec:temproal_gap}
Unlike digital environments, timing is a critical feature of real-world applications. The physical world is irreversible: once a moment has passed, it cannot be revisited. Moreover, real-world conditions can change abruptly, while LLMs cannot directly observe changes that have already occurred.

\paragraph{Scheduled actions without timely outcomes}
SimuHome requires agents to decide when to use tools in an environment that continues to change while they observe and reason. For example, agents may need to allow a device to operate for a specified duration, reassess environmental conditions after an intervention, or cancel a scheduled action when it becomes unsafe. These tasks require agents to coordinate immediate actions, future commitments, and observations as the environment changes.
Across the 15 tasks containing temporal evaluation criteria, Claude Code × Opus 5 satisfy all temporal criteria on 10 tasks (66.7\%), compared with 8 tasks (53.3\%) for both Codex × Astra/Sol. Different harnesses influence the results: Pi and Hermes are worse than Claude for Opus 5.  

Failure cases reveal several challenges. Codex and Claude Code configurations frequently submit scheduled workflows but still fail to produce the required device state at subsequent checkpoints or bring environmental conditions into the target range by the deadline. Hermes additionally exhibits cases in which required future workflows are never established;  Pi exhibits both unmet temporal requirements and model-request errors that interrupt execution. These patterns show that time-dependent tool use involves more than specifying a future timestamp: agents must establish executable actions, manage pending workflows, and verify that interventions produce the required effects within the available time. 

\paragraph{Selecting efficient time slots for health time series}

Agents can autonomously select observation windows to focus on temporally informative evidence.  We compared a fixed window of 9-hour. the autonomous Codex × GPT-5.6 Sol condition used an average of 6.4 future hours, compared with the forced 9.0-hour contiguous window, while achieving higher Exact P@1 (56.7\% vs. 53.3\%) and higher Exact P@5 (55.8\% vs. 48.6\%). Thus, forcing a fixed 9-hour observation period increased observation cost by 40.6\% without improving diagnostic accuracy. These results suggest that adaptive, agent-selected observation timing can avoid unnecessary future context and yield a more favorable performance–efficiency trade-off.

Claude Opus 5 configurations generally use substantially longer future observation windows (31–48 hours) and can achieve stronger Exact P@1 performance, with Factory × Opus 5 reaching 73.3\% using 34.6 hours on average. In contrast, Codex-based GPT-5.6 agents typically rely on only 6–10 hours of future observation while maintaining competitive accuracy: Codex × Sol under autonomous timing achieves 56.7\% Exact P@1 with 6.4 hours, and Factory × Sol reaches 60.0\% with 11.9 hours. Importantly, additional observation does not consistently improve performance—Pi × Opus 5 uses 48 hours but attains only 55.6\% Exact P@1—indicating that the trade-off is shaped by both the model and the harness, rather than observation duration alone.

\begin{table}[t]
  \centering

  \captionsetup{
    width=\textwidth,
    justification=raggedright,
    singlelinecheck=false
  }
  \caption{Ablation study of contiguous-window in Health Time-Series Analysis. Compared with the main setting, the contiguous-window is fixed as 9 hours (denoted as {``Fixed 9h''}), or as fixed 9h minus Autonomous (denoted as ``$\Delta$'').
  Values are percentages except future observation hours.}
  \label{tab:ablation-9h}

  \resizebox{\textwidth}{!}{%
  \small
  \begin{tabular}{lrrrrrrrr}
  \toprule
  Method
  & Exact P@1
  & Exact P@3
  & Exact R@3
  & Exact MRR
  & Cat. P@1
  & Cat. P@3
  & Cat. R@3
  & Future h \\
  \midrule
  Autonomous
  & 56.67 & 56.11 & 13.88 & 73.33
  & 83.33 & 72.78 & 19.29
  & 6.40 \\

  Fixed 9h
  & 53.33 & 54.44 & 14.32 & 70.00
  & 83.33 & 70.00 & 19.31
  & 9.00 \\

  $\Delta$ (Fixed 9h $-$ Autonomous)
  & -3.33
  & -1.67
  & +0.44
  & -3.33
  & +0.00
  & -2.78
  & +0.02
  & +2.60\\
  \bottomrule
  \end{tabular}
  }
\end{table}

\begin{wrapfigure}{r}{0.5\textwidth}
    \centering
    \captionsetup{
        width=0.8\linewidth,
        justification=justified,
        singlelinecheck=false
      }
    \includegraphics[width=0.45\textwidth]{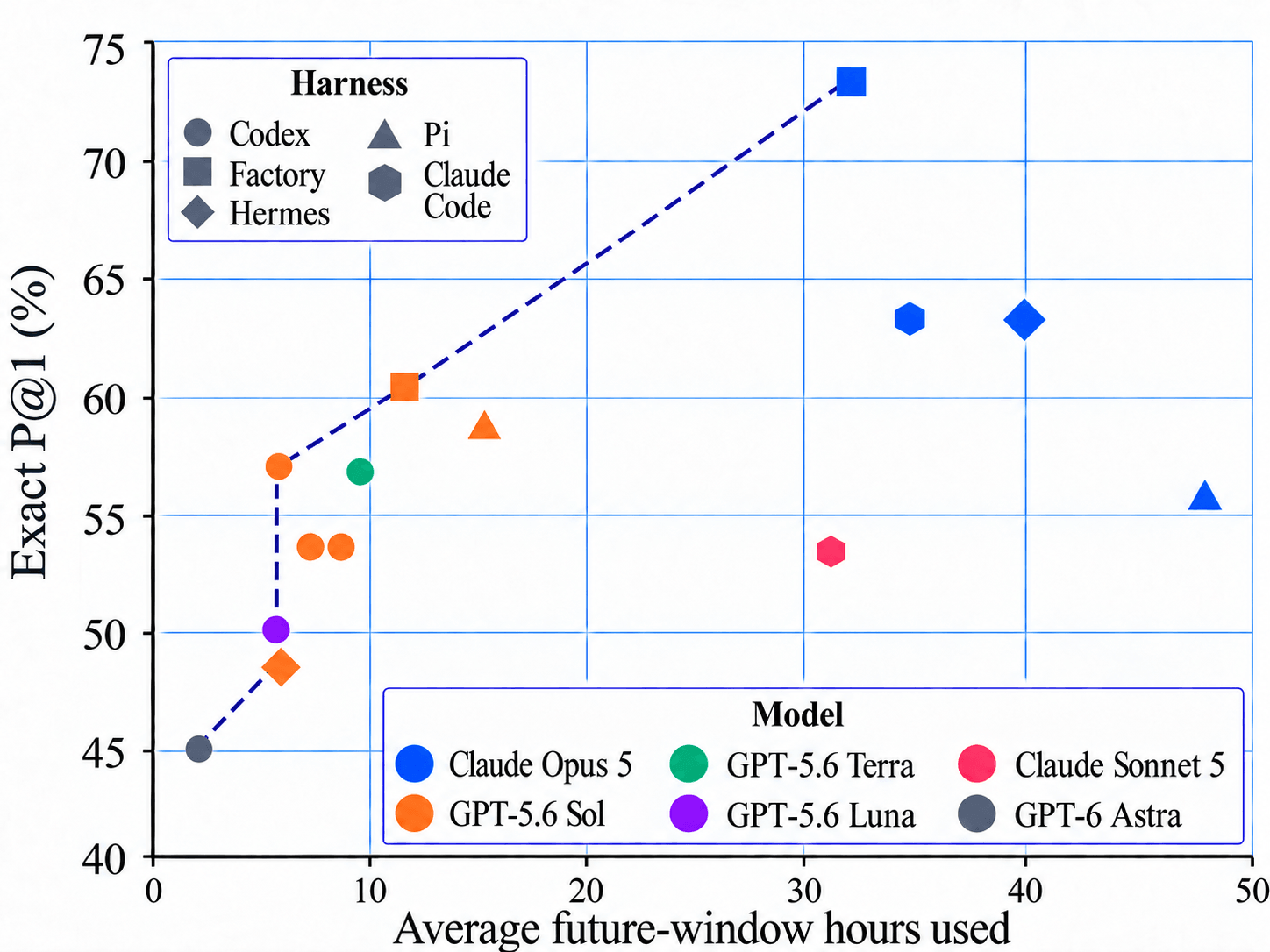}
    \caption{Performance versus Average Future-window Hours Used}
    \label{fig:token_efficiency}
    \vspace{-0.8em}
\end{wrapfigure}

\subsubsection{Multiple object control in physical environment}
\label{subsub_sec:strategy_impact}
To interact reliably with physical environments, LLMs must coordinate multiple devices and adapt to unexpected changes. Failures in these capabilities can lead to collisions, coordination errors, and delayed or unsuccessful recovery from physical anomalies.
\paragraph{Collaboration for two arms interaction}
We evaluate the multiple devices by understanding dual-arm coordination in two settings. In the Automated Biology Laboratory setting, the two arms have predefined and complementary roles: one arm performs pipetting operations, while the other moves labware between instruments. In the Functional LEGO and LEGO Building settings, the arms are not assigned fixed roles. Instead, the LLM is told that either arm can attach or detach LEGO blocks through grasping, positioning, pressing, and pulling actions. We categorize the observed failures into three types in Table~\ref{tab:dual_arm_failure_distribution}. A collision avoidance failure occurs when the two arms interfere while operating in the same workspace. Insufficient coordination occurs when one arm fails to stabilize, align, or support an object while the other performs the primary manipulation. Sequence misalignment occurs when the arms execute their assigned operations in the wrong order, resulting in an incomplete handoff between stages.
\begin{table}[t]
\centering

\captionsetup{
  width=\textwidth,
  justification=raggedright,
  singlelinecheck=false
}
\caption{Dual-arm usage and failure distribution by LLM.
Failure-type percentages are calculated over failed dual-arm cases;
the overall failure rate is calculated over all dual-arm cases;
and the dual-arm invocation rate is calculated over all evaluated cases.}
\label{tab:dual_arm_failure_distribution}

\resizebox{\textwidth}{!}{
\begin{tabular}{lccccc}
\toprule
\textbf{LLM} &
\textbf{Dual-Arm Invocation Rate} &
\textbf{Collision Avoidance} &
\textbf{Insufficient Coordination} &
\textbf{Sequence Misalignment} &
\textbf{Overall Failure Rate} \\
\midrule
GPT-6
& 30.0\% (6/20)
& 100\% (2/2)
& 0\% (0/2)
& 0\% (0/2)
& 33.3\% (2/6) \\

Claude Opus 5
& 30.0\% (9/30)
& 0\% (0/5)
& 80.0\% (4/5)
& 20.0\% (1/5)
& 55.6\% (5/9) \\

\midrule
Overall
& 30.0\% (15/50)
& 28.6\% (2/7)
& 57.1\% (4/7)
& 14.3\% (1/7)
& 46.7\% (7/15) \\
\bottomrule
\end{tabular}
}
\end{table}

LLMs with harness system can autonomously operate multiple physical devices. However, multi-device planning remains constrained by physical time, space, and action order. GPT-6 failures mainly arose from collision avoidance, whereas Claude Opus 5 failures were dominated by insufficient coordination and sequence misalignment. This indicates that GPT-6 struggled with simultaneous arm motion, while Claude Opus 5 often failed to use the two arms complementarily.

\begin{table*}[t]
\centering

\captionsetup{
  width=0.85\textwidth,
  justification=raggedright,
  singlelinecheck=false
}
\caption{Case-level response classification for dynamic physical anomalies in LEGO assembly.}
\label{tab:lego_recovery}

\small
\setlength{\tabcolsep}{5pt}
\renewcommand{\arraystretch}{1.15}

\resizebox{0.85\textwidth}{!}{%
\begin{tabularx}{\textwidth}{
    @{}
    >{\raggedright\arraybackslash}p{0.12\textwidth}
    *{3}{>{\raggedright\arraybackslash}X}
    @{}
}
\toprule
\textbf{Configuration}
& \textbf{Successful recovery}
& \textbf{Unsuccessful or incomplete recovery}
& \textbf{Undetected or misinterpreted anomaly} \\
\midrule

Codex + GPT-6
&
\textbf{40\% (4/10 cases).}
\newline
A support brick was displaced by 50\,mm and fell off the platform.
The model retrieved it, adjusted the support spacing, and then
resumed assembly.
&
\textbf{10\% (1/10 cases).}
\newline
The lower structure drifted by 4.34\,mm. The model performed a
positional correction but subsequently continued using an outdated
estimate as the drift persisted.
&
\textbf{10\% (1/10 cases).}
\newline
The two-brick structure drifted approximately 21\,mm, but the model
treated this as minor settling and reused the previous coordinates
without adjustment.
\\
\midrule

Claude Code + Opus 5
&
\textbf{50\% (5/10 cases).}
\newline
After inadvertently lifting two assembled bricks by about 30.5 mm, the model checked another viewpoint, lowered them, freed the trapped gripper, and withdrew without lifting them again.
&
\textbf{40\% (4/10 cases).}
\newline
The base brick was displaced and inverted by $180^\circ$.
The model retrieved and repositioned it but failed to correct its
orientation, then continued assembling additional layers on the
inverted base.
&
\textbf{20\% (2/10 cases).}
\newline
The supporting bricks were displaced by approximately 66--73\,mm.
The model nevertheless declared recovery successful and continued
planning placement at the previous support location.
\\
\bottomrule
\end{tabularx}%
}
\end{table*}

\paragraph{Reflections on environmental changes}

In real-world environments, many objects are not directly controlled by external devices; instead, their motion is governed by physical dynamics. The LEGO design task provides a concrete example. As discussed in~\Cref{subsub_sec:3D_gap}, millimeter-scale misalignments can prevent LEGO bricks from interlocking securely, causing them to slip and fall under gravity. We use this scenario to assess whether LLMs can detect such physical anomalies and provide effective feedback to support recovery.

We evaluate models including GPT-6 and Opus-5 and categorize the outcomes into three groups: successful recovery, unsuccessful recovery following anomaly detection, and failure to detect the anomaly. The results, reported in~\Cref{tab:lego_recovery}, show that some anomalies remain undetected, leading to missed opportunities for timely intervention. Even when anomalies are detected, recovery performance depends on the specific spatial configuration. For example, Opus-5 fails to recover when a brick is flipped by $180^\circ$, suggesting limitations in spatial reasoning.
\section{Related Work}
\label{sec:related-work}

\textbf{Harnessed LLMs in digital environments.} LLM evaluation has moved beyond isolated response generation toward task completion in harnesses that manage context, provide tools, and incorporate feedback from intermediate actions \citep{sweagent2024yang,harnessengineering2026barbaste,from2026guo}. AgentBench and GAIA assess multi-step reasoning, decision making, tool use, and information retrieval across diverse tasks \citep{agentbench2024liu,gaia2023mialon}. WebArena and OSWorld extend this evaluation to functional websites and desktop environments, where agents interact with GUIs, files, and applications \citep{webarena2024zhou,osworld2024xie}. Beyond bounded benchmark tasks, long-running harnesses have supported larger projects, including a Rust-based C compiler capable of compiling the Linux kernel, end-to-end machine-learning studies in The AI Scientist, and algorithm discovery and optimization with AlphaEvolve \citep{carlini2026compiler,aiscientist2024lu,alphaevolve2025novikov}. This work shows that task completion depends on the surrounding execution loop as well as the model. Yet existing evaluations focus largely on digital environments, leaving open how well a general-purpose harness works when its tools control external devices and return heterogeneous, physically grounded feedback.

\textbf{LLMs in physical environments.} Physical-world evaluation introduces observations and state transitions shaped by sensing conditions, device behavior, and physical constraints. Existing benchmarks address these challenges at several levels. Fixed-evidence benchmarks test physical reasoning from supplied text, images, or videos \citep{physbench2025chow,macgyver2024tian}, while active-perception benchmarks require systems to gather missing evidence by changing viewpoints or probing a scene \citep{openeqa2024majumdar,scene2025sripada,objviewbench2026pan,probe2026bhat}. Interactive benchmarks go further, requiring planning and execution as physical or device states evolve. EmbodiedBench and RoboCerebra study embodied perception, long-horizon planning, and robotic manipulation \citep{embodiedbench2025yang,robocerebra2025han}; MuBlE, BuildArena, and autonomous-laboratory systems examine construction, manipulation, and experimental execution \citep{muble2025nazarczuk,buildarena2026xia,autonomous2023boiko}; and SimuHome and DevicesWorld evaluate temporally dependent device operation and cross-device workflows \citep{simuhome2026seo,devicesworld2026li}. However, these evaluations typically focus on a bare model, an end-to-end policy, or a planner and controller tailored to a particular setting \citep{towards2025salimpour}. Even when they use an agent loop, its design is usually tied to that setting. Our benchmark instead evaluates complete model--harness systems across 11 device-grounded task families using a shared execution and evaluation pipeline, with every task involving external devices or device-generated signals.
\section{Conclusion}
We introduced W2W-Bench to evaluate complete model--harness systems across 11 interactive task families and four core capabilities. The results show that performance depends on both the model and its harness, with no configuration leading across every task. Persistent failures in spatial precision, temporal reasoning, device coordination, and recovery reveal how difficult it remains to turn language-based plans into reliable physical actions. These findings point to better observation, verification, and feedback-driven correction as priorities for future interactive harness systems.

\section*{AI Policy}
In this work, we used generative AI tools to generate synthetic datasets and assist with translation. We also used them to create or modify scientific figures or images, suggest experimental parameters, create or edit software code, create research artifacts, draft parts of the paper, improve its readability, and identify relevant literature.
We did not use generative AI tools to develop theoretical models or conceptual frameworks, formulate or prove mathematical claims, propose or refine hypotheses, design or provide feedback on research methods or experiments, implement methods, clean or reformat datasets, analyze qualitative data, interpret results, formulate survey or interview questions, transcribe research recordings, review existing literature, identify research topics or gaps, brainstorm, search for information, format references, suggest the paper’s structure, or propose a title or keywords. Some of these activities were not applicable to this work.
We reviewed all AI-assisted work. Two authors verified the accuracy of AI-generated code, figures, and draft text and tested the code where applicable. We take responsibility for the final content of this work, including any text, claims, or artifacts produced with the aid of generative AI.

\bibliographystyle{plainnat}
\bibliography{ref}

\begin{thebibliography}{33}
\providecommand{\natexlab}[1]{#1}
\providecommand{\url}[1]{\texttt{#1}}
\expandafter\ifx\csname urlstyle\endcsname\relax
  \providecommand{\doi}[1]{doi: #1}\else
  \providecommand{\doi}{doi: \begingroup \urlstyle{rm}\Url}\fi

\bibitem[Barbaste et~al.(2026)Barbaste, Darrigol, Vu, and Wiltberger]{harnessengineering2026barbaste}
Paul Barbaste, Tristan Darrigol, Germain Vu, and Tom Wiltberger.
\newblock Harness engineering: Anatomy, architecture, and evolution of coding agents--a source-code study of eleven systems.
\newblock \emph{arXiv preprint arXiv:2609.00006}, 2026.

\bibitem[Bhat et~al.(2026)Bhat, Chen, Zook, Yang, Birchfield, Blukis, and Tremblay]{probe2026bhat}
Vineet Bhat, Siyi Chen, Alex Zook, Xuning Yang, Stan Birchfield, Valts Blukis, and Jonathan Tremblay.
\newblock Probe: Manipulation-grounded visual question answering with {VLM} agents.
\newblock \emph{arXiv preprint arXiv:2608.17129}, 2026.

\bibitem[Boiko et~al.(2023)Boiko, MacKnight, Kline, and Gomes]{autonomous2023boiko}
Daniil~A. Boiko, Robert MacKnight, Ben Kline, and Gabe Gomes.
\newblock Autonomous chemical research with large language models.
\newblock \emph{Nature}, 624:\penalty0 570--578, 2023.

\bibitem[Carlini(2026)]{carlini2026compiler}
Nicholas Carlini.
\newblock {Building a C compiler with a team of parallel Claudes}.
\newblock Anthropic Engineering, February 2026.
\newblock URL \url{https://www.anthropic.com/engineering/building-c-compiler}.

\bibitem[Chow et~al.(2025)Chow, Mao, Li, Seita, Campagnolo~Guizilini, and Wang]{physbench2025chow}
Wei Chow, Jiageng Mao, Boyi Li, Daniel Seita, Vitor Campagnolo~Guizilini, and Yue Wang.
\newblock Physbench: Benchmarking and enhancing vision-language models for physical world understanding.
\newblock In \emph{International Conference on Learning Representations}, volume 2025, pages 97959--98108, 2025.

\bibitem[Collins et~al.(2022)Collins, Goel, Deng, Luthra, Xu, Gundogdu, Zhang, Vicente, Dideriksen, Arora, Guillaumin, and Malik]{collins2022abodatasetbenchmarksrealworld}
Jasmine Collins, Shubham Goel, Kenan Deng, Achleshwar Luthra, Leon Xu, Erhan Gundogdu, Xi~Zhang, Tomas F.~Yago Vicente, Thomas Dideriksen, Himanshu Arora, Matthieu Guillaumin, and Jitendra Malik.
\newblock Abo: Dataset and benchmarks for real-world 3d object understanding, 2022.
\newblock URL \url{https://arxiv.org/abs/2110.06199}.

\bibitem[Guo et~al.(2026)Guo, Hao, Wang, Fan, Luo, Li, Gao, Mei, Peng, Xu, et~al.]{from2026guo}
Jianyuan Guo, Zhiwei Hao, Chengcheng Wang, Cheng Fan, Tingzhang Luo, Hongguang Li, Ying Gao, Hefei Mei, Jiankun Peng, Rongjian Xu, et~al.
\newblock From question answering to task completion: A survey on agent system and harness design.
\newblock \emph{arXiv preprint arXiv:2606.20683}, 2026.

\bibitem[Han et~al.(2026)Han, Qiu, Liao, Huang, Gao, YAN, and Liu]{robocerebra2025han}
Songhao Han, Boxiang Qiu, Yue Liao, Siyuan Huang, Chen Gao, Shuicheng YAN, and Si~Liu.
\newblock Robocerebra: A large-scale benchmark for long-horizon robotic manipulation evaluation.
\newblock In \emph{The Thirty-ninth Annual Conference on Neural Information Processing Systems Datasets and Benchmarks Track}, 2026.
\newblock URL \url{https://openreview.net/forum?id=0JtNyaHbNx}.

\bibitem[Johnson et~al.(2019)Johnson, Pollard, and Mark]{PhysioNet-mimiciii-demo-1.4}
Alistair Johnson, Tom Pollard, and Roger Mark.
\newblock {MIMIC-III Clinical Database Demo}.
\newblock \emph{{PhysioNet}}, April 2019.
\newblock \doi{10.13026/C2HM2Q}.
\newblock URL \url{https://doi.org/10.13026/C2HM2Q}.
\newblock Version 1.4.

\bibitem[Lan et~al.(2026)Lan, Jiang, Wang, Xie, Zhang, Zhu, PEIHANG, Yang, Chen, Gao, et~al.]{lan2026autobio}
Zhiqian Lan, Yuxuan Jiang, Ruiqi Wang, Xuanbing Xie, Rongkui Zhang, Yicheng Zhu, LI~PEIHANG, Tianshuo Yang, Tianxing Chen, Haoyu Gao, et~al.
\newblock Autobio: A simulation and benchmark for robotic automation in digital biology laboratory.
\newblock In \emph{International Conference on Learning Representations}, volume 2026, pages 146176--146196, 2026.

\bibitem[Li et~al.(2026)Li, Geng, Wang, Li, Yang, Chen, Yao, Fan, Ren, Zhao, et~al.]{devicesworld2026li}
Huatao Li, Xinwei Geng, Yuheng Wang, Yutong Li, Runde Yang, Hantao Chen, Shu Yao, Jingru Fan, Xuhui Ren, Yuanyuan Zhao, et~al.
\newblock Devicesworld: Benchmarking cross-device agents in heterogeneous environments.
\newblock \emph{arXiv preprint arXiv:2607.13465}, 2026.

\bibitem[Liu et~al.(2024)Liu, Yu, Zhang, Xu, Lei, Lai, Gu, Ding, Men, Yang, et~al.]{agentbench2024liu}
Xiao Liu, Hao Yu, Hanchen Zhang, Yifan Xu, Xuanyu Lei, Hanyu Lai, Yu~Gu, Hangliang Ding, Kaiwen Men, Kejuan Yang, et~al.
\newblock {AgentBench}: Evaluating {LLMs} as agents.
\newblock In \emph{International Conference on Learning Representations}, volume 2024, pages 52989--53046, 2024.

\bibitem[Lu et~al.(2024)Lu, Lu, Lange, Foerster, Clune, and Ha]{aiscientist2024lu}
Chris Lu, Cong Lu, Robert~Tjarko Lange, Jakob~Nicolaus Foerster, Jeff Clune, and David Ha.
\newblock {The AI Scientist}: Towards fully automated open-ended scientific discovery.
\newblock \emph{ArXiv}, abs/2408.06292, 2024.
\newblock URL \url{https://api.semanticscholar.org/CorpusID:271854887}.

\bibitem[Majumdar et~al.(2024)Majumdar, Ajay, Zhang, Putta, Yenamandra, Henaff, Silwal, Mcvay, Maksymets, Arnaud, et~al.]{openeqa2024majumdar}
Arjun Majumdar, Anurag Ajay, Xiaohan Zhang, Pranav Putta, Sriram Yenamandra, Mikael Henaff, Sneha Silwal, Paul Mcvay, Oleksandr Maksymets, Sergio Arnaud, et~al.
\newblock Openeqa: Embodied question answering in the era of foundation models.
\newblock In \emph{IEEE/CVF Conference on Computer Vision and Pattern Recognition (CVPR)}, pages 16488--16498, 2024.

\bibitem[Mialon et~al.(2024)Mialon, Fourrier, Wolf, LeCun, and Scialom]{gaia2023mialon}
Gregoire Mialon, Clementine Fourrier, Thomas Wolf, Yann LeCun, and Thomas Scialom.
\newblock {GAIA}: A benchmark for general {AI} assistants.
\newblock In \emph{International Conference on Learning Representations}, volume 2024, pages 9025--9049, 2024.

\bibitem[Nazarczuk et~al.(2025)Nazarczuk, Stepanova, Behrens, Hoffmann, and Mikolajczyk]{muble2025nazarczuk}
Michal Nazarczuk, Karla Stepanova, Jan~Kristof Behrens, Matej Hoffmann, and Krystian Mikolajczyk.
\newblock Muble: Mujoco and blender simulation environment and benchmark for task planning in robot manipulation.
\newblock \emph{arXiv preprint arXiv:2503.02834}, 2025.

\bibitem[Novikov et~al.(2025)Novikov, Vu, Eisenberger, Dupont, Huang, Wagner, Shirobokov, Kozlovskii, Ruiz, Mehrabian, Kumar, See, Chaudhuri, Holland, Davies, Nowozin, Kohli, and Balog]{alphaevolve2025novikov}
Alexander Novikov, Ng{\^a}n Vu, Marvin Eisenberger, Emilien Dupont, Po-Sen Huang, Adam~Zsolt Wagner, Sergey Shirobokov, Borislav~M. Kozlovskii, Francisco J.~R. Ruiz, Abbas Mehrabian, M.~Pawan Kumar, Abigail See, Swarat Chaudhuri, George Holland, Alex Davies, Sebastian Nowozin, Pushmeet Kohli, and Matej Balog.
\newblock {AlphaEvolve}: A coding agent for scientific and algorithmic discovery.
\newblock \emph{ArXiv}, abs/2506.13131, 2025.
\newblock URL \url{https://api.semanticscholar.org/CorpusID:278658695}.

\bibitem[{OpenAI}(2026)]{openai2026navierstokes}
{OpenAI}.
\newblock {On the Navier--Stokes Millennium Prize Problem}, September 2026.
\newblock URL \url{https://openai.com/index/navier-stokes-solution/}.

\bibitem[Pan et~al.(2026)Pan, Hu, Huang, Wingender, and Bennewitz]{objviewbench2026pan}
Sicong Pan, Hao Hu, Xuying Huang, Benno Wingender, and Maren Bennewitz.
\newblock Objview-bench: Rethinking difficulty and deployment for object-centric view planning.
\newblock \emph{arXiv preprint arXiv:2605.10707}, 2026.

\bibitem[Pollard et~al.(2026)Pollard, Moody, Lehman, Gow, Fernandes, Xie, Johnson, Mark, and Heldt]{pollard_physionet_2026}
Tom Pollard, Benjamin~E. Moody, Li-wei~H. Lehman, Brian~J. Gow, Chrystinne Fernandes, Chen Xie, Alistair Johnson, Roger~G. Mark, and Thomas Heldt.
\newblock {PhysioNet} as a global platform for biomedical research.
\newblock \emph{Nature Health}, 1\penalty0 (8):\penalty0 792--795, August 2026.
\newblock ISSN 3005-0693.
\newblock \doi{10.1038/s44360-026-00096-z}.
\newblock URL \url{https://doi.org/10.1038/s44360-026-00096-z}.

\bibitem[Salimpour et~al.(2025)Salimpour, Fu, Rachwal, Bertrand, O'Sullivan, Jakob, Keramat, Militano, Toffetti, Edelman, and Pena~Queralta]{towards2025salimpour}
Sahar Salimpour, Lei Fu, Kajetan Rachwal, Pascal Bertrand, Kevin O'Sullivan, Robert Jakob, Farhad Keramat, Leonardo Militano, Giovanni Toffetti, Harry Edelman, and Jorge Pena~Queralta.
\newblock Towards embodied agentic {AI}: Review and classification of {LLM}- and {VLM}-driven robot autonomy and interaction.
\newblock \emph{arXiv preprint arXiv:2508.05294}, 2025.

\bibitem[Seo et~al.(2026)Seo, Yang, Pyo, Kim, Lee, and Jo]{simuhome2026seo}
Gyuhyeon Seo, Jungwoo Yang, Junseong Pyo, Nalim Kim, Jonggeun Lee, and Yohan Jo.
\newblock Simuhome: A temporal- and environment-aware benchmark for smart home {LLM} agents.
\newblock In \emph{The Fourteenth International Conference on Learning Representations}, 2026.
\newblock URL \url{https://openreview.net/forum?id=LCS1WsGvha}.

\bibitem[Sripada et~al.(2025)Sripada, Carter, Guerin, and Ghalamzan]{scene2025sripada}
Venkatesh Sripada, Samuel Carter, Frank Guerin, and Amir Ghalamzan.
\newblock Scene exploration by vision-language models.
\newblock \emph{arXiv preprint arXiv:2409.17641}, 2025.

\bibitem[Tian et~al.(2024)Tian, Ravichander, Qin, Le~Bras, Marjieh, Peng, Choi, Griffiths, and Brahman]{macgyver2024tian}
Yufei Tian, Abhilasha Ravichander, Lianhui Qin, Ronan Le~Bras, Raja Marjieh, Nanyun Peng, Yejin Choi, Thomas~L. Griffiths, and Faeze Brahman.
\newblock Macgyver: Are large language models creative problem solvers?
\newblock In \emph{Conference of the North American Chapter of the Association for Computational Linguistics (NAACL)}, 2024.

\bibitem[Tito et~al.(2023{\natexlab{a}})Tito, Karatzas, and Valveny]{tito2023hierarchical}
Ruben Tito, Dimosthenis Karatzas, and Ernest Valveny.
\newblock Hierarchical multimodal transformers for multipage docvqa.
\newblock \emph{Pattern Recognition}, 144:\penalty0 109834, 2023{\natexlab{a}}.

\bibitem[Tito et~al.(2023{\natexlab{b}})Tito, Karatzas, and Valveny]{tito2023hierarchicalmultimodaltransformersmultipage}
Rubèn Tito, Dimosthenis Karatzas, and Ernest Valveny.
\newblock Hierarchical multimodal transformers for multi-page docvqa, 2023{\natexlab{b}}.
\newblock URL \url{https://arxiv.org/abs/2212.05935}.

\bibitem[Wu et~al.(2015)Wu, Song, Khosla, Yu, Zhang, Tang, and Xiao]{wu20153dshapenetsdeeprepresentation}
Zhirong Wu, Shuran Song, Aditya Khosla, Fisher Yu, Linguang Zhang, Xiaoou Tang, and Jianxiong Xiao.
\newblock 3d shapenets: A deep representation for volumetric shapes, 2015.
\newblock URL \url{https://arxiv.org/abs/1406.5670}.

\bibitem[Xia et~al.(2026)Xia, Gao, Deng, Wei, Qian, Yu, and Wu]{buildarena2026xia}
Tian Xia, Tianrun Gao, Wenhao Deng, Long Wei, Xiaowei Qian, Chenglei Yu, and Tailin Wu.
\newblock Buildarena: A physics-aligned interactive benchmark of {LLMs} for engineering construction.
\newblock In \emph{International Conference on Machine Learning (ICML)}, 2026.

\bibitem[Xie et~al.(2024)Xie, Zhang, Chen, Li, Zhao, Cao, Hua, Cheng, Shin, Lei, et~al.]{osworld2024xie}
Tianbao Xie, Danyang Zhang, Jixuan Chen, Xiaochuan Li, Siheng Zhao, Ruisheng Cao, Toh~J Hua, Zhoujun Cheng, Dongchan Shin, Fangyu Lei, et~al.
\newblock {OSWorld}: Benchmarking multimodal agents for open-ended tasks in real computer environments.
\newblock \emph{Advances in Neural Information Processing Systems}, 37:\penalty0 52040--52094, 2024.

\bibitem[Yang et~al.(2024)Yang, Jimenez, Wettig, Lieret, Yao, Narasimhan, and Press]{sweagent2024yang}
John Yang, Carlos Jimenez, Alexander Wettig, Kilian Lieret, Shunyu Yao, Karthik Narasimhan, and Ofir Press.
\newblock Swe- agent: Agent-computer interfaces enable automated software engineering.
\newblock \emph{Advances in Neural Information Processing Systems}, 37:\penalty0 50528--50652, 2024.

\bibitem[Yang et~al.(2025)Yang, Chen, Zhang, Zhao, Qian, Wang, Wang, Koripella, Movahedi, Li, Ji, Zhang, and Zhang]{embodiedbench2025yang}
Rui Yang, Hanyang Chen, Junyu Zhang, Mark Zhao, Cheng Qian, Kangrui Wang, Qineng Wang, Teja~Venkat Koripella, Marziyeh Movahedi, Manling Li, Heng Ji, Huan Zhang, and Tong Zhang.
\newblock {E}mbodied{B}ench: Comprehensive benchmarking multi-modal large language models for vision-driven embodied agents.
\newblock In \emph{Proceedings of the 42nd International Conference on Machine Learning}, volume 267 of \emph{Proceedings of Machine Learning Research}, pages 70576--70631. PMLR, 2025.

\bibitem[Yao et~al.(2023)Yao, Zhao, Yu, Du, Shafran, Narasimhan, and Cao]{react2023yao}
Shunyu Yao, Jeffrey Zhao, Dian Yu, Nan Du, Izhak Shafran, Karthik~R Narasimhan, and Yuan Cao.
\newblock React: Synergizing reasoning and acting in language models.
\newblock In \emph{The Eleventh International Conference on Learning Representations}, 2023.
\newblock URL \url{https://openreview.net/forum?id=WE_vluYUL-X}.

\bibitem[Zhou et~al.(2023)Zhou, Xu, Zhu, Zhou, Lo, Sridhar, Cheng, Bisk, Fried, Alon, and Neubig]{webarena2024zhou}
Shuyan Zhou, Frank~F. Xu, Hao Zhu, Xuhui Zhou, Robert Lo, Abishek Sridhar, Xianyi Cheng, Yonatan Bisk, Daniel Fried, Uri Alon, and Graham Neubig.
\newblock Webarena: A realistic web environment for building autonomous agents.
\newblock \emph{ArXiv}, abs/2307.13854, 2023.
\newblock URL \url{https://api.semanticscholar.org/CorpusID:260164780}.

\end{thebibliography}

\clearpage
\appendix
\section{Task Pipeline}
\label{app:task}
\subsection{Document Image Capture}

\begin{enumerate}
      \item {Initialize}
      \begin{itemize}
          \item Place an identity document in a randomized 3D scene.
          \item Initialize the lighting, occlusions, and camera settings.
      \end{itemize}

      \item {Harness Loop}
      \begin{itemize}
          \item Capture an image and inspect the visual feedback.
          \item Adjust the camera pose, distance, exposure, or flash.
          \item Repeat until the image satisfies the quality requirements.
      \end{itemize}

      \item {Finalize}
      \begin{itemize}
          \item Select and submit one captured image.
          \item Evaluate its framing, sharpness, lighting, glare, shadow, and occlusion.
      \end{itemize}
  \end{enumerate}

\subsection{3D Scanning}
\begin{enumerate}
      \item {Initialize}
      \begin{itemize}
          \item Load a target object and initialize the Franka wrist RGB-D camera.
          \item Capture an initial keyframe and generate safe scan candidates.
      \end{itemize}

      \item {Harness Loop}
      \begin{itemize}
          \item Select a collision-free camera trajectory and capture RGB-D frames.
          \item Fuse the frames into the current point-cloud reconstruction.
          \item Review coverage feedback and repeat for up to eight scans.
      \end{itemize}

      \item {Finalize}
      \begin{itemize}
          \item Submit the reconstructed point cloud.
          \item Compare it with the hidden target geometry using precision, recall, and F-score.
      \end{itemize}
  \end{enumerate}

\subsection{Object Recognition}
\begin{enumerate}
      \item {Initialize}
      \begin{itemize}
          \item Load a 3D object in a scene with randomized occlusions.
          \item Initialize the camera and candidate object labels.
      \end{itemize}

      \item {Harness Loop}
      \begin{itemize}
          \item Select the camera azimuth, elevation, and distance.
          \item Capture and inspect the object from multiple viewpoints.
          \item Continue until sufficient visual evidence is obtained.
      \end{itemize}

      \item {Finalize}
      \begin{itemize}
          \item Submit the predicted object label.
          \item Evaluate label correctness and whether an informative image was captured.
      \end{itemize}
  \end{enumerate}

\subsection{Multipage Question Answering}
\begin{enumerate}
      \item {Initialize}
      \begin{itemize}
          \item Load a multi-page document and a question requiring four fields.
          \item Provide the page-viewing tools and a limited exploration budget.
      \end{itemize}

      \item {Harness Loop}
      \begin{itemize}
          \item Select and inspect document pages relevant to the requested fields.
          \item Combine evidence across pages and revisit uncertain information when needed.
          \item Continue until all fields have sufficient supporting evidence.
      \end{itemize}

      \item {Finalize}
      \begin{itemize}
          \item Submit the four extracted field values and their supporting page references.
          \item Evaluate answer correctness, evidence precision and recall, and output validity.
      \end{itemize}
  \end{enumerate}

\subsection{Health Time-Series Analysis}
The Harness follows an anchor-time active diagnosis pipeline:


\begin{enumerate}
\item{{Initialize.}}
  \begin{itemize}
      \item Load the patient's basic signals from the 24 hours before the anchor time.
      \item Summarize signal quality, trends, and abnormalities.
      \item Record an initial differential diagnosis.
  \end{itemize}

\item{{Harness Loop.}}
  \begin{itemize}
      \item Decide whether additional observation is needed.
      \item Select a future window and collect basic signals, SpO$_2$, or an ECG proxy.
      \item Reassess the diagnosis and repeat when useful, within the 72-hour horizon.
  \end{itemize}

\item{{Finalize.}}
  \begin{itemize}
      \item Review the currently available clinical context.
      \item Search the ICD-9 dictionary and submit ranked predictions.
      \item Lock the episode and perform post-hoc evaluation using ground truth.
  \end{itemize}
\end{enumerate}
  \paragraph{Note.}
  The inference agent cannot access patient-associated ICD-9 labels or discharge
  diagnoses. Ground truth is used exclusively by the post-terminal evaluation
  harness and cannot be used to revise the predictions.

\subsection{SimuHome}

\begin{enumerate}
\item{{Initialize.}}
  \begin{itemize}
      \item Initialize the simulated home, devices, workflows, and hidden task conditions.
      \item Provide the household objective and GUI-based observation tools.
  \end{itemize}

\item{{Harness Loop.}}
  \begin{itemize}
      \item Inspect relevant rooms, devices, environmental variables, and workflows.
      \item Select device-control actions based on the observed state.
      \item Observe the resulting changes and continue until the objective is satisfied.
  \end{itemize}

\item{{Finalize.}}
  \begin{itemize}
      \item Report the completed actions and final household state.
      \item Evaluate task outcome, safety constraints, runtime success, and observation evidence.
  \end{itemize}
\end{enumerate}
\subsection{3D Video Generation}

\begin{enumerate}
\item{Initialize.}
  \begin{itemize}
      \item The harness loads the product model and generates a common baseline capture.
      \item The harness presents the Detail, Hero, Wide, Functional, and Creative views.
      \item The harness exposes the camera configuration, motion constraints, and remaining interaction budget.
  \end{itemize}

\item{Execution Loop.}
  \begin{itemize}
      \item The LLM inspects the five current images and identifies presentation problems.
      \item The LLM revises the camera positions, viewing directions, focus, or exposure.
      \item The harness checks parameter limits, inverse kinematics, collisions, and transitions between views.
      \item The harness executes a feasible plan and returns five new images; the LLM then decides whether further adjustment is needed.
  \end{itemize}

\item{Finalize.}
  \begin{itemize}
      \item The LLM decides whether the latest capture is sufficient.
      \item If it's sufficient, the LLM submits the latest set of five photographs as the final product presentation.
  \end{itemize}
\end{enumerate}

\paragraph{Note.}
The LLM receives feasibility errors but no visual-quality scores during the episode.

\subsection{LEGO Building}

\begin{enumerate}
\item{Initialize.}
  \begin{itemize}
      \item The harness loads a MuJoCo workspace containing loose LEGO pieces and two robot arms.
      \item The harness loads the target structure, language construction steps, and available inventory.
      \item The LLM inspects synchronized wrist-camera images, calibration, gripper states, and robot proprioception.
  \end{itemize}

\item{Execution Loop.}
  \begin{itemize}
      \item The LLM selects the arm, grasp direction, approach path, placement pose, gripper command, and possible dual-arm support.
      \item The harness rejects actions that violate workspace, speed, joint, or reachability constraints.
      \item The harness executes one physical action and returns the resulting images, contact feedback, and robot state.
      \item The LLM verifies each grasp, placement, press, or release, then adjusts the viewpoint, retries, regrasps, or rebuilds when necessary.
  \end{itemize}

\item{Finalize.}
  \begin{itemize}
      \item The LLM releases the structure, clears the grippers, and performs a final visual inspection.
      \item The LLM submits either a completion judgment or a specific unresolved failure.
  \end{itemize}
\end{enumerate}

\subsection{Automated Biology Laboratory}

\begin{enumerate}
\item{Initialize.}
  \begin{itemize}
      \item The harness initializes a seeded experiment with liquids, wells, consumables, devices, possible faults, and resource budgets.
      \item The harness provides the public scientific objective, permitted operations, recent results, and remaining budget.
      \item The harness exposes camera-based evidence and its references without structured laboratory ground truth.
  \end{itemize}

\item{Execution Loop.}
  \begin{itemize}
      \item The LLM acquires relevant views and interprets the observed laboratory state.
      \item The LLM selects one liquid-handling, labware, or device operation at a time.
      \item The harness validates volumes, tip state, contamination constraints, observation freshness, budgets, and device order before execution.
      \item The harness returns the execution outcome, and the LLM updates the experimental plan.
      \item After an uncertain or partial effect, the LLM obtains fresh evidence and reconciles the state before retrying.
  \end{itemize}

\item{Finalize.}
  \begin{itemize}
      \item The LLM verifies whether the scientific objective has been reached.
      \item The LLM places the pipette in a safe, clean state and resolves any observable uncertainty, then ends the experiment.
  \end{itemize}
\end{enumerate}

\paragraph{Note.}
Hidden container state, fault schedules, and target predicates are unavailable to the LLM during the experiment.

\subsection{Functional LEGO Design}

\begin{enumerate}
\item{Initialize.}
  \begin{itemize}
      \item The harness presents a functional objective, fixed workspace fixtures, and a component inventory.
      \item The LLM inspects the scene, browses available components, and reads their manuals.
      \item The harness withholds any reference structure, component-role assignment, or prescribed construction sequence.
  \end{itemize}

\item{Execution Loop.}
  \begin{itemize}
      \item The LLM selects useful components and formulates a structure intended to satisfy the functional constraints.
      \item The LLM builds through grasping, transport, placement, support, removal, and correction actions.
      \item After every physical action, the harness returns the changed structure; the LLM obtains complementary views when needed.
      \item The LLM revises component choices, local placements, or the overall design according to visual and execution feedback.
  \end{itemize}

\item{Finalize.}
  \begin{itemize}
      \item The LLM releases the structure and clears the robot arms.
      \item The LLM inspects the final state and the result video for visible defects or unsupported components.
      \item The LLM submits the current physical design.
  \end{itemize}
\end{enumerate}

\paragraph{Note.}
The LLM receives no reference design or prescribed solution and must decide when the observed structure is ready for submission.

\subsection{Tabletop Games}

\begin{enumerate}
\item{Initialize.}
  \begin{itemize}
      \item The harness initializes Hanabi, Regicide, or Pandemic from a seed and synchronizes the game state with a MuJoCo tabletop.
      \item The harness assigns one seat to the tested LLM and controls the remaining seats with game-specific heuristic policies.
      \item The harness provides the LLM with its player-visible state and legal actions while preserving hidden game information.
  \end{itemize}

\item{Execution Loop.}
  \begin{itemize}
      \item The LLM selects a symbolic game action according to the current strategy.
      \item The harness checks rule legality and translates a legal choice into physical card or piece transfers.
      \item The harness checks robot reachability, inverse kinematics, joint limits, and collisions during execution.
      \item If a transfer fails, the harness restores the affected object and preserves the game state, allowing the LLM to retry.
      \item After successful execution, the harness advances the rule engine and synchronizes the board; the LLM then reconsiders the next decision from the updated state.
  \end{itemize}

\item{Finalize.}
  \begin{itemize}
      \item The interaction continues until the game reaches a terminal state or the turn limit.
      \item If three consecutive action attempts fail, the harness aborts the current game.
  \end{itemize}
\end{enumerate}

\paragraph{Note.}
A rule-legal action advances the game only after its physical transfers succeed.
The harness does not replace three unsuccessful LLM decisions with a random legal fallback.

\section{Tool List}
\label{app:tool}
\noindent
This inventory classifies task-facing interfaces by their primary role in information acquisition, environmental interaction, and execution management. It contains \textbf{61 tool entries and 16 Tabletop game-action entries}, for a total of \textbf{77 entries}. Counts are task-specific: identically named interfaces in different tasks are counted separately. 

\subsection{Category Overview}

\paragraph{Reading the detailed tables.}
Each table represents one primary category, so the category is not repeated
in every row. Task icons appear below each tool name in the first column
and use the same convention as the overview; paired
LEGO icons denote the shared interface set for LEGO Building and Functional
LEGO Design. Short labels distinguish continuous and incremental scanning,
and Hanabi, Regicide, and Pandemic. Tool descriptions summarize each
interface's operation and constraints; the last column identifies the
associated devices or information sources.

\subsection{Observation Acquisition}
Observation acquisition covers the visual observations and temporally grounded observations. It includes active camera control, experimental measurement, and selective access to recorded signals or document images. Multipage QA page and region requests acquire task evidence; they are not reference-knowledge lookups.

\begin{tooltable}{Observation Acquisition interfaces (14 entries).}{tab:sensing}
\tool{capture_card}\newline \taskicon{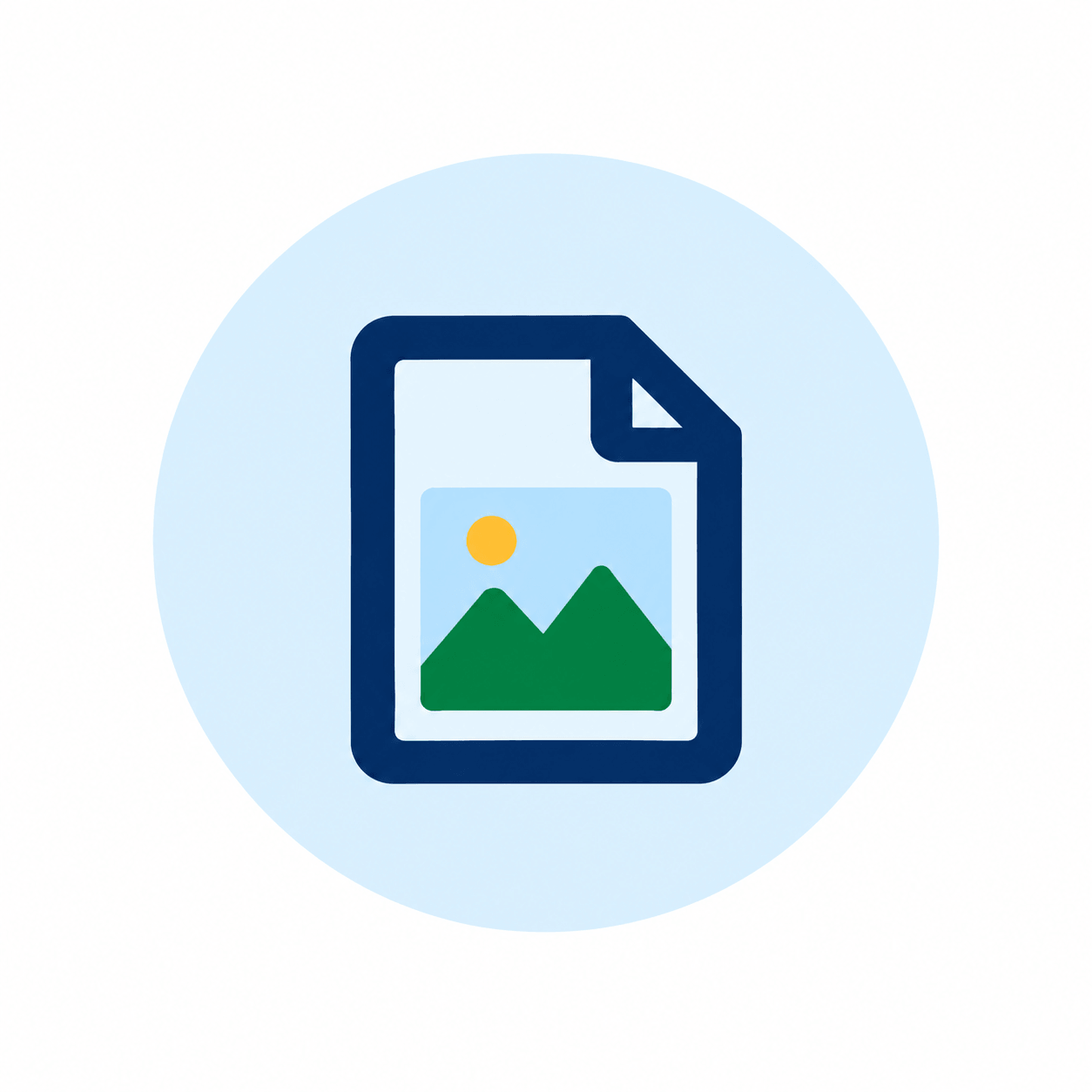} & Captures a document image with adjustable viewpoint and exposure. & Card scene and camera. \\
\addlinespace[0.35em]
\tool{capture_video_scan}\newline \taskicon{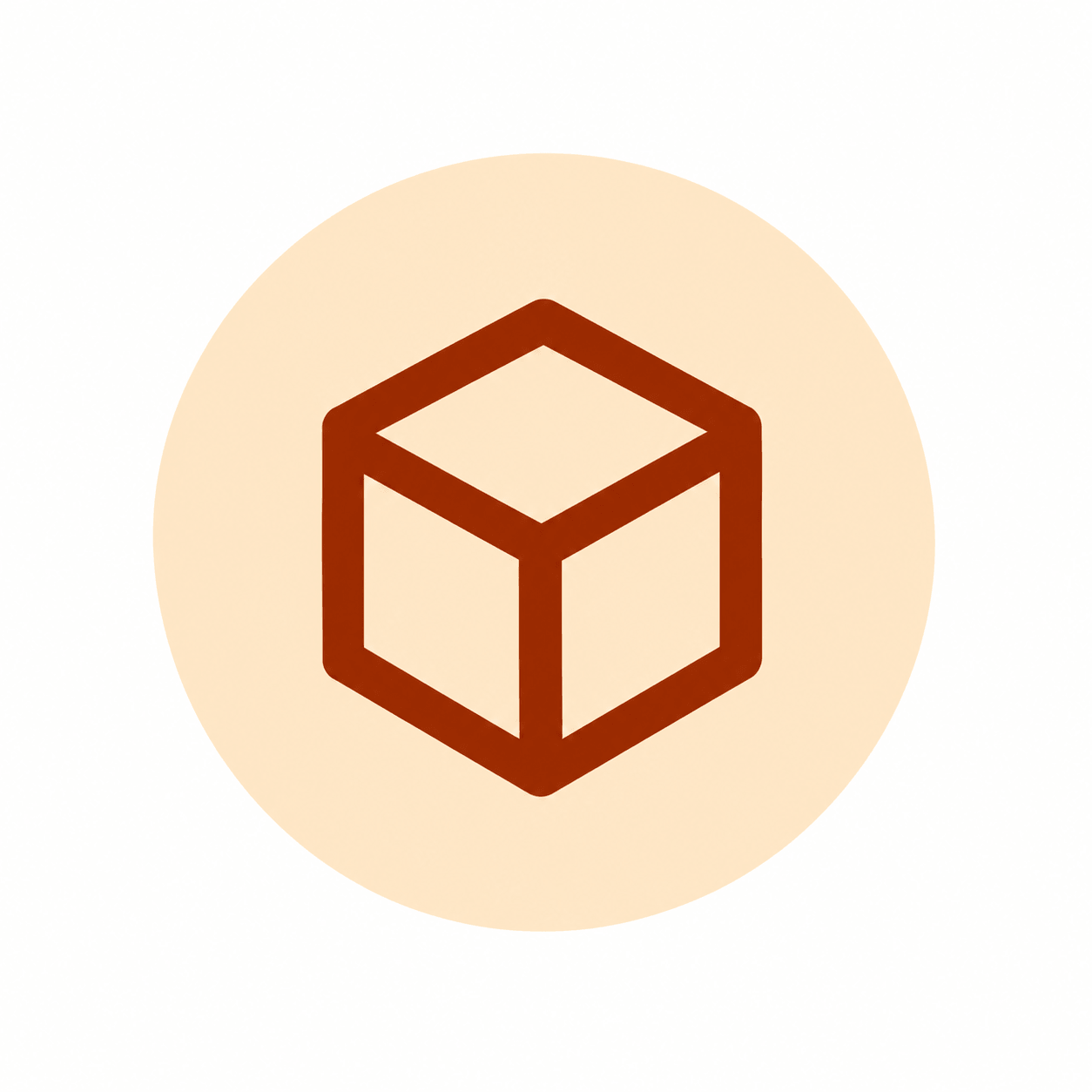}\hspace{0.25em}{\scriptsize Continuous} & Acquires and fuses multiple frames along a continuous camera trajectory. & Wrist RGB-D camera, robot arm, and reconstruction system. \\
\addlinespace[0.35em]
\tool{capture_reconstruction_view}\newline \taskicon{figure/icons/icon_3Dscan.png}\hspace{0.25em}{\scriptsize Incremental} & Acquires a single RGB-D view and fuses it into the point-cloud reconstruction. & Franka arm, RGB-D camera, and point-cloud system. \\
\addlinespace[0.35em]
\tool{capture}\newline \taskicon{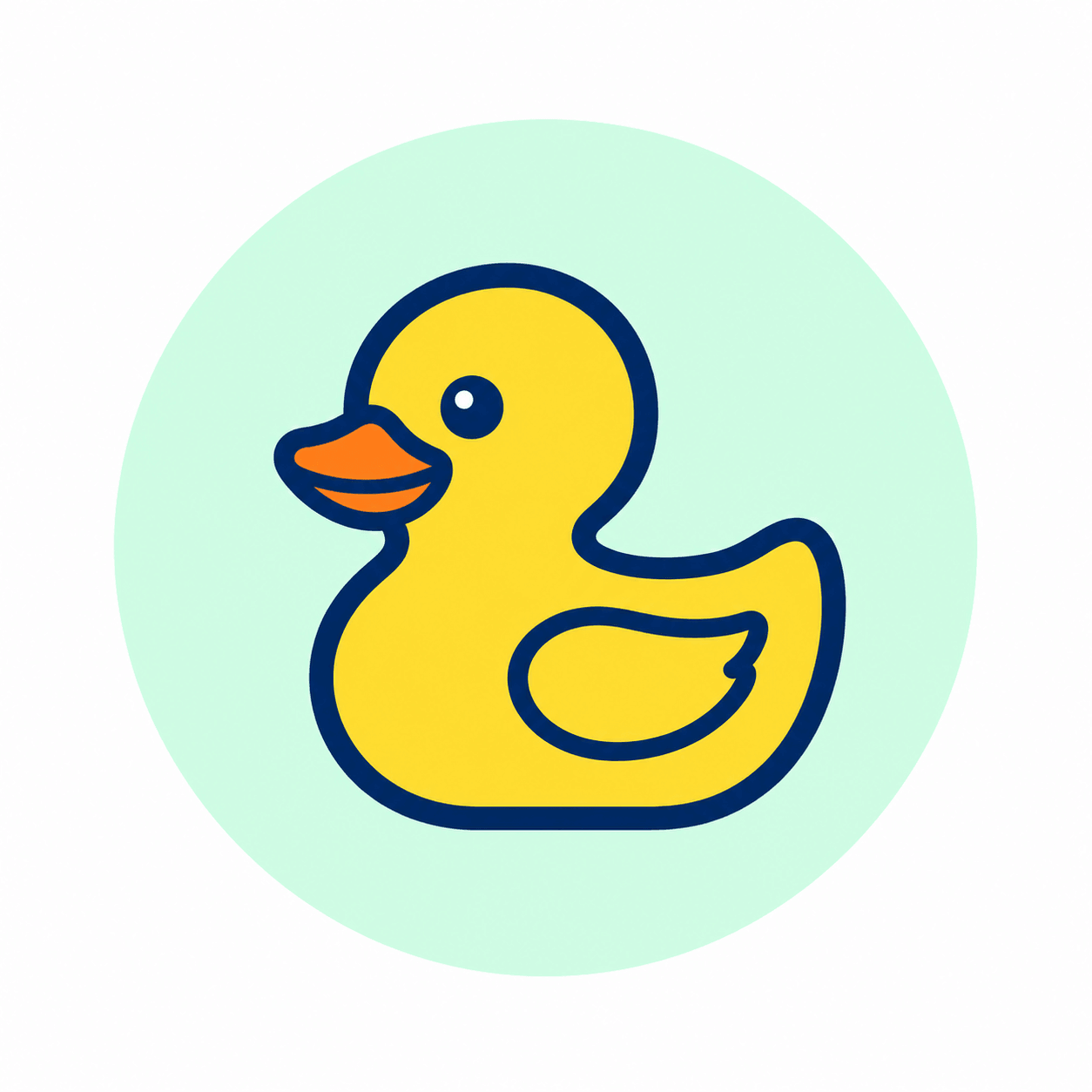} & Captures an object from a selected camera viewpoint. & Object scene and camera. \\
\addlinespace[0.35em]
\tool{request_page}\newline \taskicon{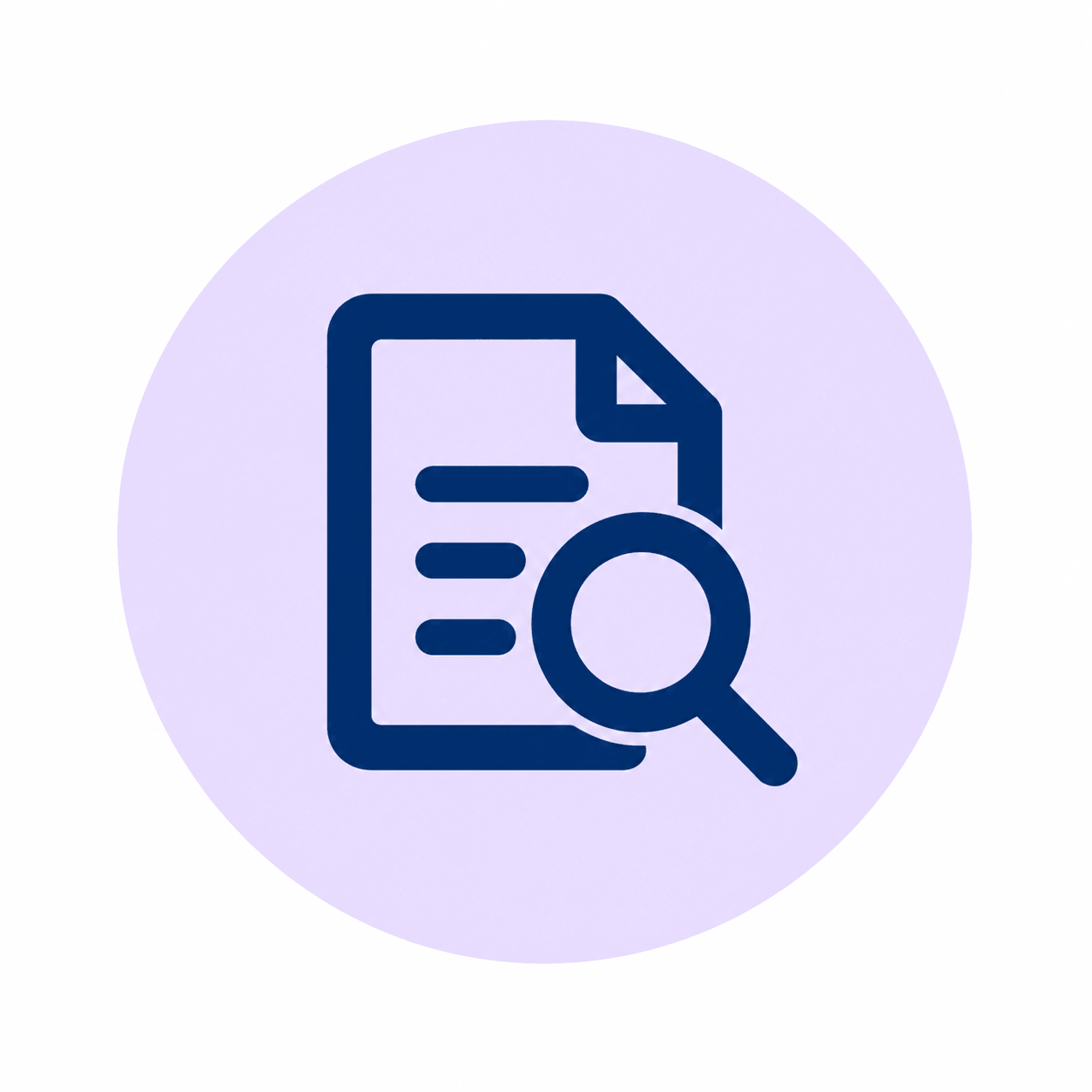} & Retrieves a selected document page for visual inspection. & A specified page of a multipage document. \\
\tool{zoom_region}\newline \taskicon{figure/icons/icon_QA.png} & Magnifies a selected region of a document page for closer inspection. & A specified rectangular region of a document page. \\
\addlinespace[0.35em]
\tool{get_historical_basic_signals}\newline \taskicon{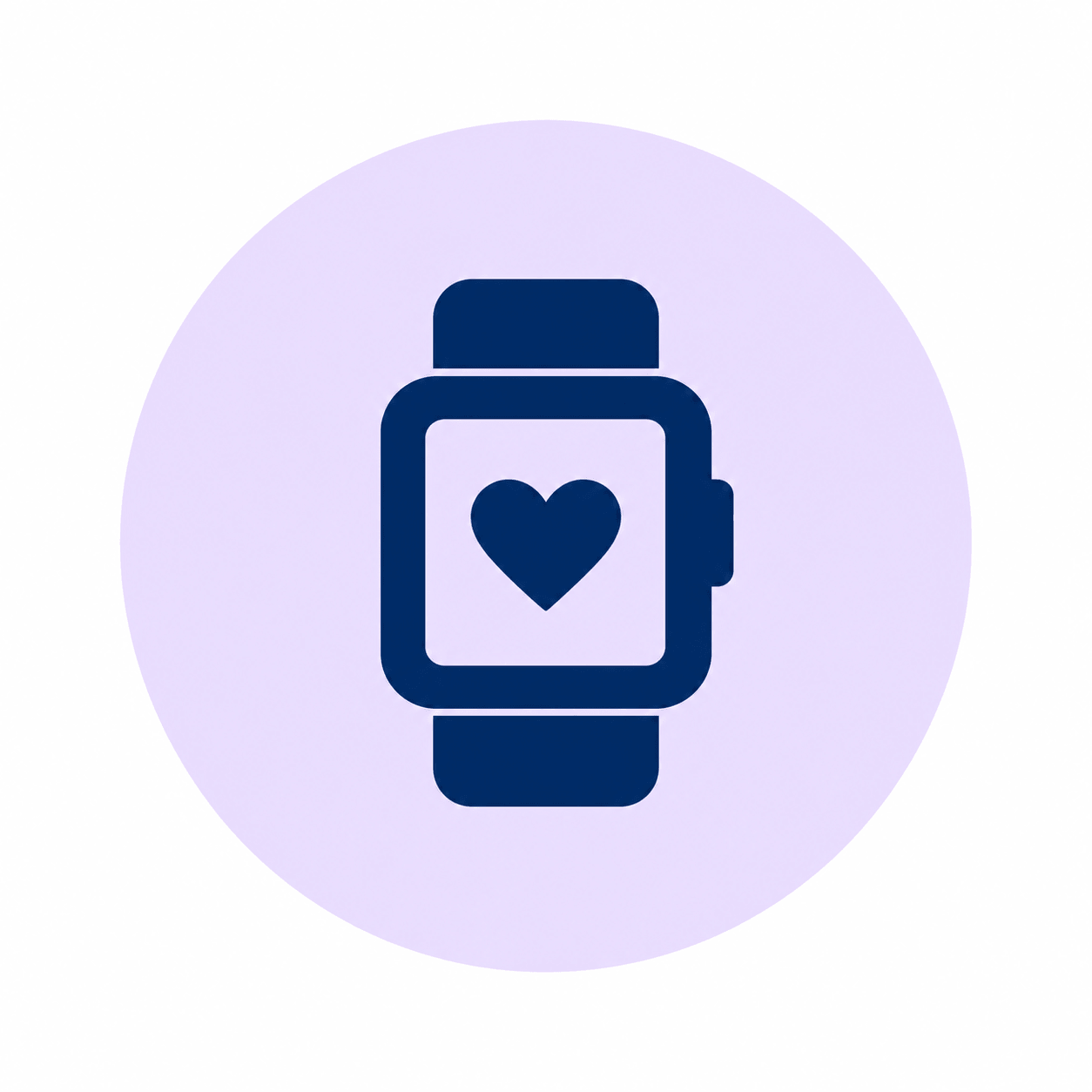} & Retrieves pre-anchor physiological observations and patient demographics. & Pre-anchor patient records and physiological signals. \\
\tool{observe_basic_window}\newline \taskicon{figure/icons/icon_health.png} & Reveals multivariate physiological signals within an observation window without advancing time. & Heart rate, temperature, respiratory rate, and blood pressure. \\
\tool{collect_spo2}\newline \taskicon{figure/icons/icon_health.png} & Retrieves oxygen-saturation observations with coverage and trend statistics. & SpO\textsubscript{2} data. \\
\tool{collect_ecg_summary}\newline \taskicon{figure/icons/icon_health.png} & Returns a heart-rate variation summary; does not analyze ECG waveforms or morphology. & Heart-rate variation proxy data; no ECG morphology. \\
\addlinespace[0.35em]
\tool{capture}\newline \taskicon{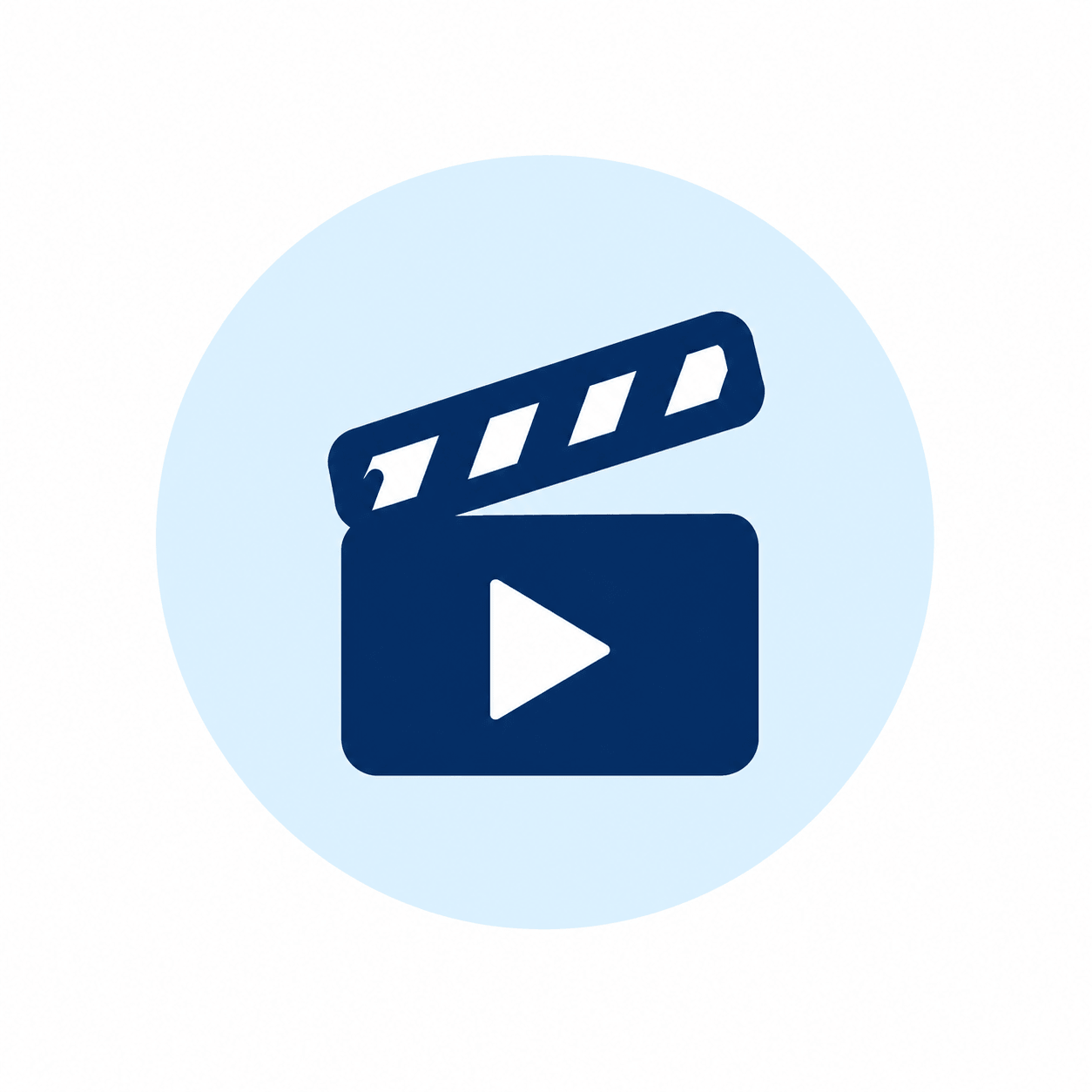} & Moves the mobile base and arm to acquire the five-shot photo sequence. & Mobile base, Franka arm, and camera. \\
\addlinespace[0.35em]
\tool{observe_lego}\newline \taskicon{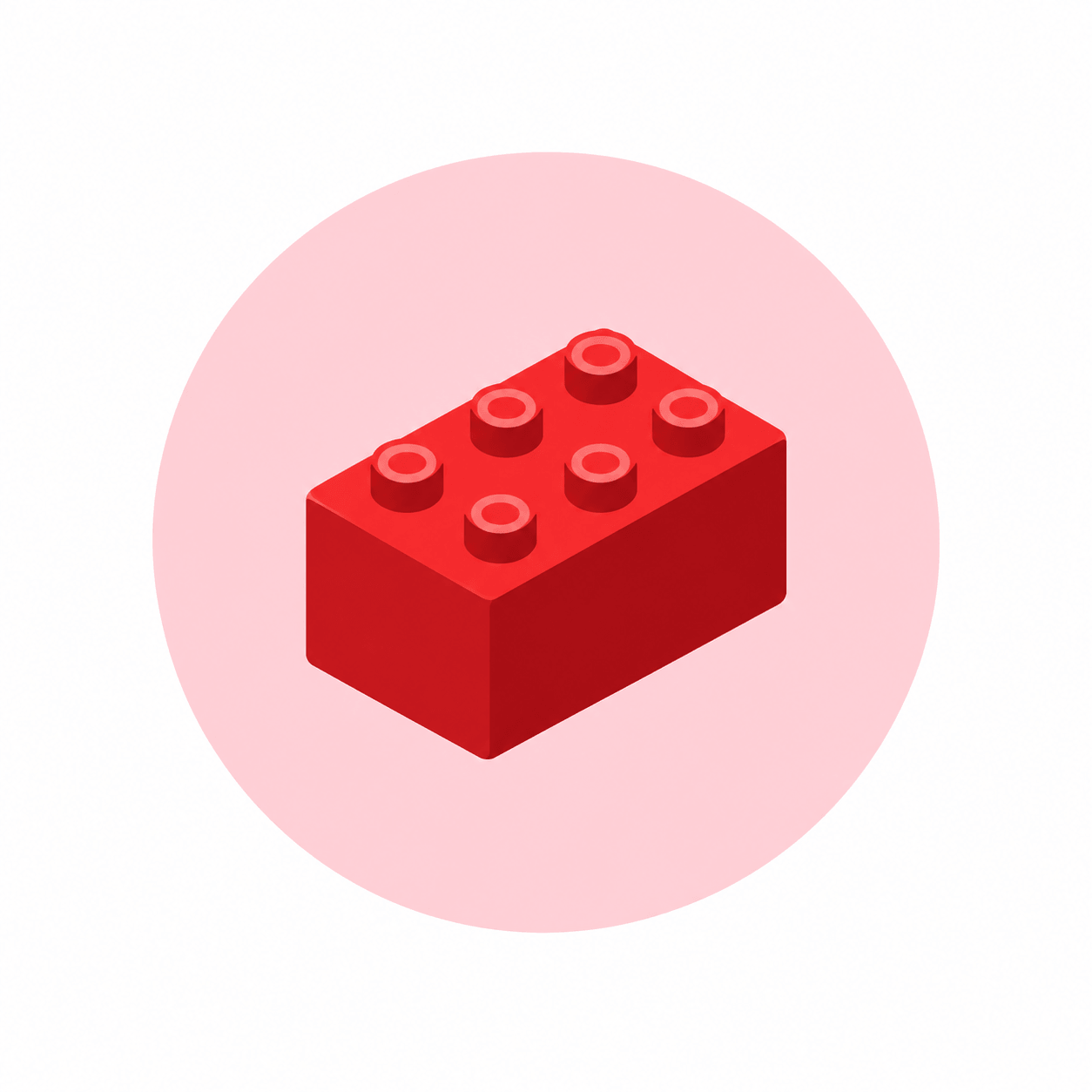}\hspace{0.15em}\taskicon{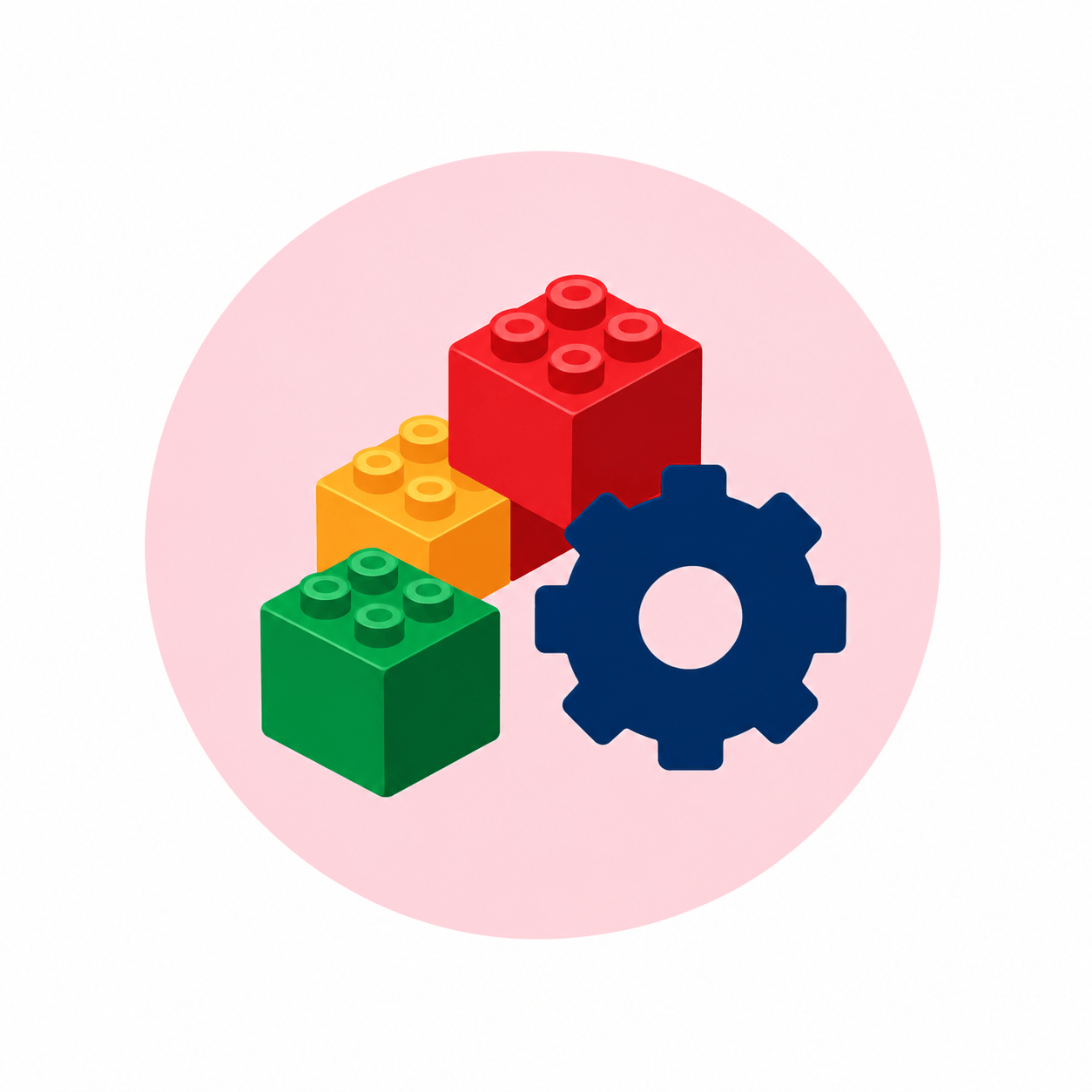} & Returns dual-wrist images, robot state, and camera calibration information. & Dual-wrist RGB cameras and robot state. \\
\addlinespace[0.35em]
\tool{view_camera}\newline \taskicon{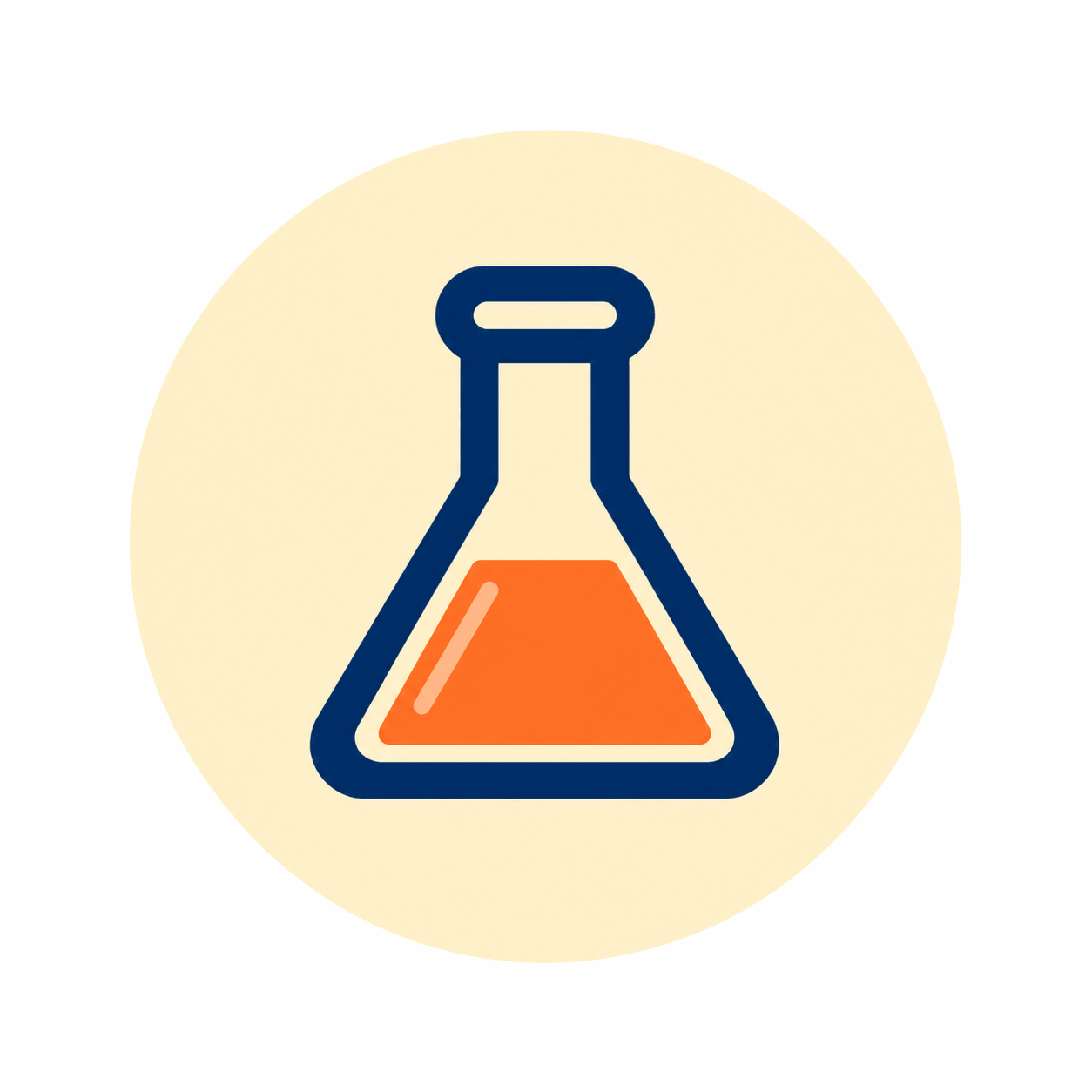} & Returns a rendered laboratory view with its state version. & Rendered camera view of the laboratory environment. \\
\tool{read_plate}\newline \taskicon{figure/icons/icon_lab.png} & Operates the plate reader and produces an experimental measurement artifact. & Plate reader and experimental plate. \\
\end{tooltable}

\subsection{Action Execution}
These interfaces execute selected actions through device commands or domain-specific operations. All 16 Tabletop game actions belong here, including passing and yielding: their effects are defined by game rules. Tabletop requests are interpreted by the harness rather than exposing direct robot pose control.

\begin{tooltable}{Action Execution interfaces (28 entries).}{tab:actuation}
\tool{execute_command}\newline \taskicon{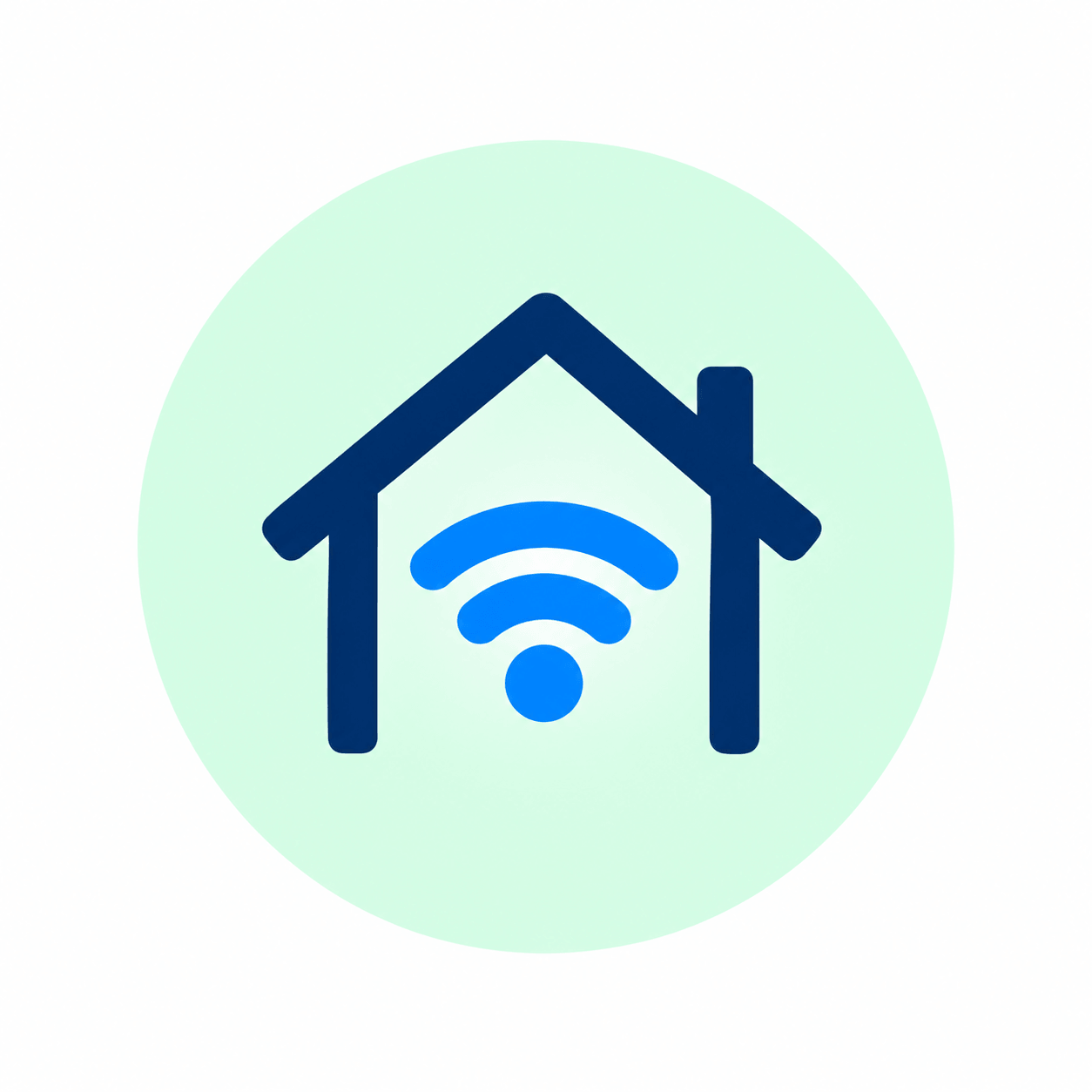} & Executes device commands to control the environment. & Matter-style device interfaces. \\
\tool{write_attribute}\newline \taskicon{figure/icons/icon_simu.png} & Changes writable device attributes and configuration values. & Writable attributes of smart-home devices. \\
\addlinespace[0.35em]
\tool{act_lego}\newline \taskicon{figure/icons/icon_lego1.png}\hspace{0.15em}\taskicon{figure/icons/icon_lego2.png} & Controls both robot arms through tool-center-point poses, yaw, and gripper commands. & Two robot arms, grippers, and LEGO parts. \\
\addlinespace[0.35em]
\tool{home_robot}\newline \taskicon{figure/icons/icon_lab.png} & Moves the laboratory robot to its home configuration. & Laboratory robot. \\
\tool{pick_up_tip}\newline \taskicon{figure/icons/icon_lab.png} & Loads a sterile pipette tip for liquid handling. & Pipetting robot and sterile tips. \\
\tool{drop_tip}\newline \taskicon{figure/icons/icon_lab.png} & Disposes of the attached pipette tip. & Pipetting robot and waste area. \\
\tool{aspirate}\newline \taskicon{figure/icons/icon_lab.png} & Aspirates a specified liquid volume from a source container. & Pipette and source container. \\
\tool{dispense}\newline \taskicon{figure/icons/icon_lab.png} & Dispenses a specified liquid volume into a destination container. & Pipette and destination container. \\
\tool{mix}\newline \taskicon{figure/icons/icon_lab.png} & Mixes liquid in a container using the pipette. & Pipette and container. \\
\tool{move_labware}\newline \taskicon{figure/icons/icon_lab.png} & Moves labware between registered laboratory slots. & Laboratory robot, labware, and registered slots. \\
\tool{operate_shaker}\newline \taskicon{figure/icons/icon_lab.png} & Runs a shaking program on the laboratory shaker. & Laboratory shaker. \\
\tool{incubate}\newline \taskicon{figure/icons/icon_lab.png} & Performs timed incubation with temperature control. & Incubation equipment and experimental samples. \\
\addlinespace[0.35em]
\tool{play}\newline \taskicon{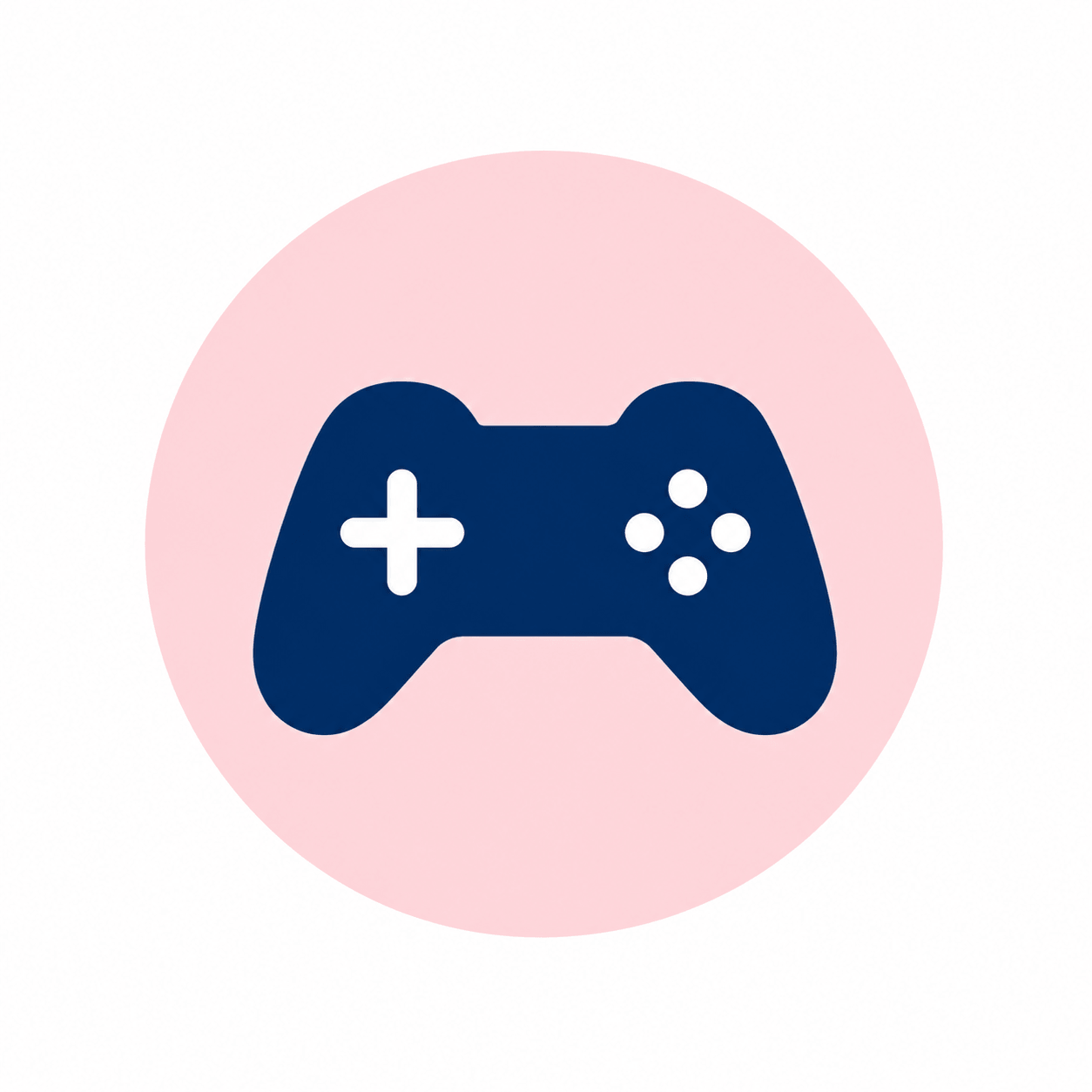}\hspace{0.25em}{\scriptsize Hanabi} & Plays a card to advance a firework pile, with a failure penalty for an invalid play. & Hand cards, fireworks area, game engine, and robot execution layer. \\
\tool{discard}\newline \taskicon{figure/icons/icon_game.png}\hspace{0.25em}{\scriptsize Hanabi} & Discards a card to recover a hint token. & Hand cards, discard area, game engine, and robot execution layer. \\
\tool{hint}\newline \taskicon{figure/icons/icon_game.png}\hspace{0.25em}{\scriptsize Hanabi} & Provides a teammate with a color or rank hint about their cards. & Color or rank hints about a teammate's cards. \\
\addlinespace[0.35em]
\tool{play}\newline \taskicon{figure/icons/icon_game.png}\hspace{0.25em}{\scriptsize Regicide} & Attacks the enemy using a card or card combination. & Hand cards, enemy state, game engine, and robot execution layer. \\
\tool{discard}\newline \taskicon{figure/icons/icon_game.png}\hspace{0.25em}{\scriptsize Regicide} & Discards cards to absorb damage during defense resolution. & Hand cards, discard area, game engine, and robot execution layer. \\
\tool{yield}\newline \taskicon{figure/icons/icon_game.png}\hspace{0.25em}{\scriptsize Regicide} & Skips the attack and triggers enemy-attack resolution. & Game-turn and combat-resolution state. \\
\addlinespace[0.35em]
\tool{drive}\newline \taskicon{figure/icons/icon_game.png}\hspace{0.25em}{\scriptsize Pandemic} & Moves the player pawn to an adjacent city. & City map, player pawn, and robot execution layer. \\
\tool{direct_flight}\newline \taskicon{figure/icons/icon_game.png}\hspace{0.25em}{\scriptsize Pandemic} & Moves to a city by spending its destination card. & City cards, player pawn, and robot execution layer. \\
\tool{charter_flight}\newline \taskicon{figure/icons/icon_game.png}\hspace{0.25em}{\scriptsize Pandemic} & Moves to a destination by spending the current city card. & City cards, player pawn, and robot execution layer. \\
\tool{shuttle}\newline \taskicon{figure/icons/icon_game.png}\hspace{0.25em}{\scriptsize Pandemic} & Moves between cities with research stations. & Research stations, player pawn, and robot execution layer. \\
\tool{build}\newline \taskicon{figure/icons/icon_game.png}\hspace{0.25em}{\scriptsize Pandemic} & Builds a research station in the current city. & City, station marker, and robot execution layer. \\
\tool{treat}\newline \taskicon{figure/icons/icon_game.png}\hspace{0.25em}{\scriptsize Pandemic} & Removes disease cubes from the current city. & Disease cubes, city, and robot execution layer. \\
\tool{share_give}\newline \taskicon{figure/icons/icon_game.png}\hspace{0.25em}{\scriptsize Pandemic} & Gives a city card to a co-located player. & Co-located players, city cards, and robot execution layer. \\
\tool{share_take}\newline \taskicon{figure/icons/icon_game.png}\hspace{0.25em}{\scriptsize Pandemic} & Receives a city card from a co-located player. & Co-located players, city cards, and robot execution layer. \\
\tool{cure}\newline \taskicon{figure/icons/icon_game.png}\hspace{0.25em}{\scriptsize Pandemic} & Spends city cards at a research station to discover a cure. & City cards, research station, cure state, and game engine. \\
\tool{pass}\newline \taskicon{figure/icons/icon_game.png}\hspace{0.25em}{\scriptsize Pandemic} & Ends the current player's action phase early. & Current player's turn state. \\
\end{tooltable}

\subsection{State and Feedback Inspection}
These interfaces expose current state, execution outcomes, and feedback for checking progress and intended effects. This category includes existing-image retrieval, asynchronous result collection, and visual outcome inspection. It supports verification without implying access to hidden evaluation scores. SimuHome state pages also provide environmental observations; their primary role here is state inspection.

\begin{tooltable}{State and Feedback Inspection interfaces (10 entries).}{tab:monitoring}
\tool{inspect_video_reconstruction_progress}\newline \taskicon{figure/icons/icon_3Dscan.png}\hspace{0.25em}{\scriptsize Continuous} & Reports scan history, online reconstruction metrics, contact sheets, and safe candidate scans. & Reconstruction state, contact sheets, and candidate scans. \\
\addlinespace[0.35em]
\tool{inspect_reconstruction_progress}\newline \taskicon{figure/icons/icon_3Dscan.png}\hspace{0.25em}{\scriptsize Incremental} & Reports the current image, trajectory, online reconstruction metrics, and safe candidate viewpoints. & Reconstruction state, current RGB image, and candidate viewpoints. \\
\addlinespace[0.35em]
\tool{view_home_page}\newline \taskicon{figure/icons/icon_simu.png} & Displays rooms, environmental variables, device summaries, and virtual time. & Rooms, environmental variables, and device summaries. \\
\tool{view_device_page}\newline \taskicon{figure/icons/icon_simu.png} & Displays device attributes, writable values, clusters, and supported commands. & Device attributes, clusters, writable values, and supported commands. \\
\tool{view_workflows_page}\newline \taskicon{figure/icons/icon_simu.png} & Displays scheduled workflows, execution times, steps, and status. & Workflow IDs, times, steps, and status. \\
\addlinespace[0.35em]
\tool{get_state}\newline \taskicon{figure/icons/icon_3Dvideo.png} & Reports current camera settings and the remaining interaction budget. & Current camera settings and remaining budget. \\
\tool{view_photos}\newline \taskicon{figure/icons/icon_3Dvideo.png} & Retrieves the five most recent annotated photographs without capturing new images. & The five most recent annotated photographs. \\
\addlinespace[0.35em]
\tool{observe_result_video}\newline \taskicon{figure/icons/icon_lego1.png}\hspace{0.15em}\taskicon{figure/icons/icon_lego2.png} & Inspects the assembly's behavior over time through an orbiting-camera video. & Orbiting camera and the assembly's behavior over time. \\
\tool{wait_lego}\newline \taskicon{figure/icons/icon_lego1.png}\hspace{0.15em}\taskicon{figure/icons/icon_lego2.png} & Waits for previously submitted actions and retrieves their results and images without re-execution. & Results and images from previously submitted actions. \\
\tool{lego_status}\newline \taskicon{figure/icons/icon_lego1.png}\hspace{0.15em}\taskicon{figure/icons/icon_lego2.png} & Reports running jobs and jobs with uncollected results. & IDs of running jobs or jobs with uncollected results. \\
\end{tooltable}

\subsection{Context and Resource Access}
These interfaces supply visible context, reference material, and available resources for reasoning and decision making. Physiological observations and document-page evidence belong to observation acquisition; clinical narrative context, coding dictionaries, manuals, rules, and resource catalogs belong here.

\begin{tooltable}{Context and Resource Access interfaces (7 entries).}{tab:knowledge}
\tool{list_video_reconstruction_targets}\newline \taskicon{figure/icons/icon_3Dscan.png}\hspace{0.25em}{\scriptsize Continuous} & Lists available reconstruction targets without exposing reference models. & Available reconstruction targets, excluding reference models. \\
\addlinespace[0.35em]
\tool{list_reconstruction_targets}\newline \taskicon{figure/icons/icon_3Dscan.png}\hspace{0.25em}{\scriptsize Incremental} & Lists available reconstruction target identifiers. & Available reconstruction target IDs. \\
\addlinespace[0.35em]
\tool{get_clinical_context}\newline \taskicon{figure/icons/icon_health.png} & Retrieves clinical context available before the temporal cutoff, including complaints, admission type, and started medications. & Chief complaint, admission type, and medications already started. \\
\tool{search_icd9_codes}\newline \taskicon{figure/icons/icon_health.png} & Searches the global ICD-9 dictionary without exposing patient-diagnosis associations. & Global ICD-9 dictionary without patient-diagnosis associations. \\
\addlinespace[0.35em]
\tool{view_environment_rules_page}\newline \taskicon{figure/icons/icon_simu.png} & Retrieves rules linking device actions to changes in environmental variables. & Rules linking device actions to temperature, humidity, illumination, and air quality. \\
\addlinespace[0.35em]
\tool{list_parts}\newline \taskicon{figure/icons/icon_lego1.png}\hspace{0.15em}\taskicon{figure/icons/icon_lego2.png} & Lists available LEGO part types, quantities, and colors. & Part types, quantities, and colors. \\
\tool{read_part_manual}\newline \taskicon{figure/icons/icon_lego1.png}\hspace{0.15em}\taskicon{figure/icons/icon_lego2.png} & Retrieves the generic English manual for a selected LEGO part. & Generic English manual for the specified part. \\
\end{tooltable}

\subsection{Planning and Tool Orchestration}
These interfaces support task initialization, observation planning, parameter validation, and tool orchestration. Workflow dispatch remains here because it coordinates an action sequence; individual device commands belong to action execution. This is an interface-role category, not a claim that planning is performed by a tool or that every planning-related action belongs here.

\begin{tooltable}{Planning and Tool Orchestration interfaces (8 entries).}{tab:planning}
\tool{start_video_reconstruction_episode}\newline \taskicon{figure/icons/icon_3Dscan.png}\hspace{0.25em}{\scriptsize Continuous} & Initializes a reconstruction episode and acquires the fixed initial keyframe. & Target configuration, wrist camera, and fixed initial keyframe. \\
\addlinespace[0.35em]
\tool{start_reconstruction_episode}\newline \taskicon{figure/icons/icon_3Dscan.png}\hspace{0.25em}{\scriptsize Incremental} & Initializes a reconstruction episode at the fixed initial viewpoint. & Target configuration, RGB-D camera, and fixed initial viewpoint. \\
\addlinespace[0.35em]
\tool{plan_future_window}\newline \taskicon{figure/icons/icon_health.png} & Plans a future observation window within the 72-hour post-anchor horizon. & Episode state and the 72-hour post-anchor horizon. \\
\tool{advance_time}\newline \taskicon{figure/icons/icon_health.png} & Commits the observation window and advances the simulation clock. & Current observation window and simulation clock. \\
\addlinespace[0.35em]
\tool{schedule_workflow}\newline \taskicon{figure/icons/icon_simu.png} & Schedules an action sequence at an absolute virtual time. & Workflow scheduler and absolute virtual time. \\
\tool{cancel_workflow}\newline \taskicon{figure/icons/icon_simu.png} & Cancels a workflow that has not yet executed. & Workflows that have not yet executed. \\
\tool{run_workflow_now}\newline \taskicon{figure/icons/icon_simu.png} & Executes a workflow immediately and sequentially, without automatic rollback. & Workflow executor and device-action sequence. \\
\addlinespace[0.35em]
\tool{check_plan}\newline \taskicon{figure/icons/icon_3Dvideo.png} & Validates five-shot parameters, inverse kinematics, and collision constraints. & Five-shot plan, robot kinematics, and collision model. \\
\end{tooltable}

\subsection{Assessment and Submission}
This category combines the two assessment and evidence recording interfaces with the eight submission and termination interfaces in Appendix B. Intermediate assessments make the agent's interpretation explicit; final submissions provide answers or artifacts and may terminate the episode. These agent-facing operations remain distinct from the independent evaluator described in Section 2.3.

\begin{tooltable}{Assessment and Submission interfaces (10 entries).}{tab:submission}
\tool{submit_photo}\newline \taskicon{figure/icons/icon_doc.png} & Selects a captured image for submission and triggers hidden-quality evaluation. & Specified image ID and captured photographs. \\
\addlinespace[0.35em]
\tool{finalize_video_reconstruction_episode}\newline \taskicon{figure/icons/icon_3Dscan.png}\hspace{0.25em}{\scriptsize Continuous} & Terminates reconstruction using online evidence without revealing ground-truth scores. & Current reconstruction and online evidence; no ground-truth score disclosure. \\
\addlinespace[0.35em]
\tool{finalize_reconstruction_episode}\newline \taskicon{figure/icons/icon_3Dscan.png}\hspace{0.25em}{\scriptsize Incremental} & Finalizes reconstruction with a stop reason, confidence, and output artifacts. & Reconstruction artifacts and the agent's stopping judgment. \\
\addlinespace[0.35em]
\tool{guess_label}\newline \taskicon{figure/icons/icon_object.png} & Submits the predicted object class. & Predicted object class. \\
\addlinespace[0.35em]
\tool{submit_answer}\newline \taskicon{figure/icons/icon_QA.png} & Submits four answer fields and their supporting document pages. & Four answer fields and supporting document pages. \\
\addlinespace[0.35em]
\tool{record_assessment}\newline \taskicon{figure/icons/icon_health.png} & Records an intermediate differential diagnosis, uncertainty, information gain, and proposed next step. & Agent assessment, acquired evidence, and episode record. \\
\tool{finalize_episode}\newline \taskicon{figure/icons/icon_health.png} & Validates ranked ICD-9 predictions and the final assessment, then atomically locks the result. & Ranked ICD-9 predictions, final assessment, and retrieved codes. \\
\addlinespace[0.35em]
\tool{finish}\newline \taskicon{figure/icons/icon_3Dvideo.png} & Submits the latest five photographs and permanently terminates the task. & The latest five photographs. \\
\addlinespace[0.35em]
\tool{finish_lego}\newline \taskicon{figure/icons/icon_lego1.png}\hspace{0.15em}\taskicon{figure/icons/icon_lego2.png} & Submits the agent's visual completion judgment and terminates the task. & Assembly observations and the agent's completion judgment. \\
\addlinespace[0.35em]
\tool{report_perception}\newline \taskicon{figure/icons/icon_lab.png} & Records the agent's scene interpretation with citations to camera evidence. & Camera observations and the agent's perception report. \\
\end{tooltable}

\paragraph{Classification convention.}
Each interface has one primary category for aggregation; auxiliary tags retain
overlapping capabilities. The categories do not form mutually exclusive
implementation properties or a one-to-one mapping to loop stages. Capture
may entail robot motion, state inspection may reveal observations, and
orchestration may execute device actions. Classification follows the principal
agent-facing role. Agent reasoning and strategy updates are not counted as
tools unless an explicit interface exposes or records them. Hidden ground
truth and independent evaluation remain outside the agent-facing inventory.

\section{Dataset}
\label{app_sec:dataset}
\paragraph{\emph{Document Image Capture.}} The dataset contains 20 synthetic samples spanning four fictional identity-document types. Document content and SVG textures are generated programmatically and placed in MuJoCo tabletop scenes with varied layouts and lighting. Occlusions, shadows, and glare introduce capture challenges without using real identity documents.

\paragraph{\emph{3D Scanning.}} A total of 30 samples are drawn from ModelNet10\cite{wu20153dshapenetsdeeprepresentation} and ABO20\cite{collins2022abodatasetbenchmarksrealworld}. The 3D assets serve as reference objects for simulated scanning, where observations are acquired from agent-selected viewpoints and accumulated for reconstruction. Reconstruction feedback informs subsequent view selection.

\paragraph{\emph{Object Recognition.}} Eight public glTF assets—Avocado, Fish, Boombox, Duck, Water Bottle, Lantern, Antique Camera, and Helmet—form the basis of a 20-sample synthetic dataset. These assets are reused across scenes with different rotations, lighting conditions, and occlusions. Camera azimuth, elevation, and distance can be adjusted during an episode to obtain additional views of the same object.

\paragraph{\emph{Multipage Question Answering.}} Evaluation data are curated from the validation split of lmms-lab-encoder/MP-DocVQA\cite{tito2023hierarchicalmultimodaltransformersmultipage}. An initial pool of 30 long documents is filtered into 20 evaluation samples, each comprising 20 pages and four questions, for a total of 80 question-answer fields. Documents are presented through page and region inspection, with supporting evidence distributed across one to four pages.

\paragraph{\emph{Health Time-Series Analysis.}} MIMIC-III Clinical Database Demo v1.4\cite{PhysioNet-mimiciii-demo-1.4, pollard_physionet_2026} supplies de-identified clinical records from 100 patients. Physiological measurements are presented as incomplete, noisy time-series observations, with additional measurements or later observation windows exposed on request. The dataset’s patient count is distinct from the number of constructed evaluation episodes.
\paragraph{\emph{SimuHome.}} Rather than loading fixed sample files, SimuHome V2 generates episodes from 36 task templates and deterministic seeds. A shared home configuration contains six room entries, 39 devices, and 17 device types. Initialization assigns device settings, environmental conditions, virtual time, schedules, and workflows, while episode configurations introduce failures, unavailable devices, safety locks, or conflicting objectives. The simulator renders these states through a browser interface and records actions, state changes, and evaluation results.
\paragraph{\emph{3D Video Generation.}} ABO20\cite{collins2022abodatasetbenchmarksrealworld} provides 20 product samples used as renderable 3D assets. Each asset is observed through rendered views controlled by camera radius, height, elevation, and focus~\footnote{https://jimmccarthy.org/wp-content/uploads/2016/06/checklist-fiveshot-0.3-small.pdf}. These views support iterative camera adjustment and the production of a final 360-degree product video.

\paragraph{\emph{LEGO Building.}} Basic LEGO-20 comprises 20 samples with predefined target structures. Each target is instantiated as an assembly task in a simulated environment, where piece placements produce updated physical states and wrist-camera observations. The resulting interaction captures intermediate configurations, including pose errors, failed connections, and collisions.

\paragraph{\emph{Automated Biology Laboratory.}} AutoBio~\cite{lan2026autobio} is organized around 10 drug-screening task templates, which are combined with seeds to generate episodes in the Fast Twin laboratory simulator. Generation assigns target wells, source concentrations and volumes, tip inventories, plate orientations, and equipment states, then introduces conditions such as occlusion, missing consumables, drift, delays, or partial action effects. The simulator maintains liquid and device states and produces camera observations, GUI feedback, and synthetic experimental readings.

\paragraph{\emph{Tabletop Games.}} Data for Hanabi~\footnote{https://boardgamegeek.com/boardgame/98778/hanabi}, Regicide~\footnote{https://www.regicidegame.com/}, and Pandemic~\footnote{https://boardgamegeek.com/boardgame/30549/pandemic} are generated from game rules and random seeds. Initialization shuffles the relevant decks and establishes player states, along with the city map and infection states for Pandemic. Logical states are mapped to tabletop object positions, and legal game actions are converted into robotic movements checked for reachability and collisions in MuJoCo. Only successful executions update the game state; screenshots, serialized states, and JSON records are saved as episode outputs or caches.

\paragraph{\emph{Functional LEGO Design.}} Open Functional LEGO-20 contains 20 samples defined by functional objectives and geometric constraints. Each sample combines its objective with a library of digital LEGO components, allowing multiple candidate designs. Component arrangements are inspected through design feedback for stability, connectivity, and clearance, and revised configurations form the interaction trajectory.

\section{Metrics of Tasks}
\label{app:metric}

This appendix reports the component-level, strict-verdict, and process metrics used to diagnose performance beyond the primary metrics in Section~\ref{subsec:evaluation_setting}. Unless stated otherwise, these metrics are averaged over episodes and are not used to rank systems in the main-results table.

\paragraph{\taskicon{figure/icons/icon_doc.png}Document Image Capture.}
\label{app:metric:docs}
The objective is to produce a document image that is geometrically and photometrically well formed. For episode $i$, the evaluator measures seven quality dimensions---\emph{glare}, \emph{shadow}, \emph{perspective distortion}, \emph{corner visibility}, \emph{sharpness}, \emph{framing}, and \emph{exposure}---with scores $q_{ik}^{\mathrm{docs}}\in[0,1]$. Let $\tau_k$ be the acceptance threshold for dimension $k$. The primary metric is the strict \emph{Complete Rate}, whose episode-level success indicator is

\begin{equation}
    b_i^{\mathrm{docs}}
    =\prod_{k=1}^{7}
     \mathbf{1}\!\left[q_{ik}^{\mathrm{docs}}\geq\tau_k\right].
\end{equation}
Thus, an episode succeeds only when all seven quality checks pass; a high score on one dimension cannot compensate for failure on another. All seven dimensions are computed by the evaluation backend from the images produced by the model's actions; no LLM judge is used.

In addition to the strict {Complete Rate}, we report the mean document-quality score
\begin{equation}
    s_i^{\mathrm{docs}}
    =\frac{1}{7}\sum_{k=1}^{7}q_{ik}^{\mathrm{docs}}.
\end{equation}
This auxiliary score summarizes average image quality but can mask a critical failure such as severe shadow, clipped corners, or unusable perspective. We also report the component success score
\begin{equation}
    s_i^{\mathrm{docs\text{-}comp}}
    =\frac{1}{7}\sum_{k=1}^{7}
     \mathbf{1}\!\left[q_{ik}^{\mathrm{docs}}\geq\tau_k\right],
\end{equation}
as well as each of the seven underlying quality scores: glare, shadow, perspective distortion, corner visibility, sharpness, framing, and exposure.

\paragraph{\taskicon{figure/icons/icon_3Dscan.png}3D Scanning.}
\label{app:metric:scan}
This task requires reconstructing a complete object surface from actively selected RGB-D views under camera-execution constraints. Let $\mathcal{R}_i$ and $\mathcal{G}_i$ denote the reconstructed and ground-truth point sets. At distance tolerance $\delta$, surface precision, recall, and F-score are
\begin{align}
    \operatorname{Prec}_i(\delta)
    &=\frac{1}{|\mathcal{R}_i|}\sum_{r\in\mathcal{R}_i}
      \mathbf{1}\!\left[\min_{g\in\mathcal{G}_i}\|r-g\|_2\leq\delta\right],\\
    \operatorname{Rec}_i(\delta)
    &=\frac{1}{|\mathcal{G}_i|}\sum_{g\in\mathcal{G}_i}
      \mathbf{1}\!\left[\min_{r\in\mathcal{R}_i}\|g-r\|_2\leq\delta\right],\\
    F_i(\delta)
    &=\frac{2\operatorname{Prec}_i(\delta)\operatorname{Rec}_i(\delta)}
      {\operatorname{Prec}_i(\delta)+\operatorname{Rec}_i(\delta)}.
\end{align}
The primary episode score is $s_i^{\mathrm{scan}}=F_i(10\,\mathrm{mm})$.

A strict reconstruction pass is defined as
\begin{equation}
    b_i^{\mathrm{scan}}
    =\mathbf{1}\!\left[F_i(10\,\mathrm{mm})\geq0.90\right],
\end{equation}
together with any required execution-validity gates in robot-based variants. Geometric diagnostics include precision and recall at $10\,\mathrm{mm}$, accuracy, completeness, Chamfer distance, and F-scores at $2$, $5$, and $20\,\mathrm{mm}$. Process diagnostics include incremental surface coverage, multi-view support, azimuth and elevation coverage, number of views or scans, camera-path length, achieved-pose error, inverse-kinematics success, collision, and minimum clearance.

\paragraph{\taskicon{figure/icons/icon_object.png}Object Recognition.}
\label{app:metric:recognition}
The model identifies an object by actively selecting informative viewpoints and analyzing the acquired partial views. An episode succeeds only when the submitted label matches the canonical ground-truth label (including accepted aliases) and at least one acquired view is informative:
\begin{equation}
    b_i^{\mathrm{recog}}
    =\mathbf{1}\!\left[\widehat{y}_i=y_i\right]
     \cdot
     \mathbf{1}\!\left[\sum_v I_{i,v}^{\mathrm{recog}}>0\right],
\end{equation}
where $I_{i,v}^{\mathrm{recog}}$ indicates whether view $v$ provides sufficiently visible and informative evidence.

View informativeness $I_{i,v}^{\mathrm{recog}}$ is evaluated from four visual evidence dimensions: visibility, target scale, framing, and image content. We report label accuracy and image informativeness separately to distinguish recognition failures from acquisition failures. View-sector coverage, use of a near view, and captures per successful episode characterize the diversity and efficiency of visual exploration.

\paragraph{\taskicon{figure/icons/icon_QA.png}Multipage Question Answering.}
\label{app:metric:qa}
Questions may require evidence distributed across multiple document pages. Let $A_i^{\mathrm{QA}}$ be the fraction of answers that match the ground truth and $E_i^{\mathrm{QA}}$ the fraction of questions for which the required evidence page was visited and correctly cited. The primary episode score is
\begin{equation}
    s_i^{\mathrm{QA}}
    =0.75A_i^{\mathrm{QA}}+0.25E_i^{\mathrm{QA}}.
\end{equation}

The strict-pass indicator requires every answer to be correct, every question to have valid supporting evidence, and no citation to refer to an unvisited page. Additional diagnostics include field-level exact accuracy, answer-string similarity, evidence precision and recall, and evidence-question accuracy.

\paragraph{\taskicon{figure/icons/icon_health.png}Health Time-Series Analysis.}
\label{app:metric:health}
The model produces a ranked ICD-9 differential diagnosis while selectively acquiring additional physiological observations. The primary metric is exact Precision@1, a binary episode-level top-1 hit:
\begin{equation}
    b_i^{\mathrm{health}}
    =\mathbf{1}\!\left[\widehat{y}_{i,1}\in\mathcal{Y}_i\right],
    \qquad
    \operatorname{ExactP@1}
    =\frac{1}{N}\sum_{i=1}^{N}b_i^{\mathrm{health}},
\end{equation}
where $\mathcal{Y}_i$ is the set of exact ground-truth ICD-9 codes.

A separate strict verdict, {verified success}, requires protocol compliance and an exact match to the primary diagnosis among the top three predictions. Exact-code and three-digit-category Precision@$K$ and Recall@$K$ for $K\in\{1,3,5\}$, mean reciprocal rank, average precision, and primary-diagnosis Hit@1/3/5 characterize ranking quality at different granularities. Observation hours, acquired windows, sensor calls, and invalid tool calls measure information-acquisition efficiency but are not included in the diagnosis score.

\paragraph{\taskicon{figure/icons/icon_simu.png}SimuHome.}
\label{app:metric:simuhome}
The objective is to achieve a household goal while satisfying its temporal, safety, recovery, and execution requirements. Each episode is decomposed into binary clauses grouped into outcome, temporal, safety, recovery, and runtime categories. Let $q_{ic}$ be the fraction of clauses passed in category $c$ when that category is present in episode $i$. The primary score is
\begin{equation}
    s_i^{\mathrm{home}}
    =\sum_{c\in\mathcal{C}_i}\widetilde{w}_{ic}q_{ic},
    \qquad
    \widetilde{w}_{ic}
    =\frac{w_c}{\sum_{c'\in\mathcal{C}_i}w_{c'}},
\end{equation}
where the base weights for \emph{outcome}, \emph{temporal}, \emph{safety}, \emph{recovery}, and \emph{runtime} are $(45,20,25,10,10)$, respectively.

The hard-pass indicator is the conjunction of every gating clause in an episode. Observation-evidence clauses are scored separately and do not block success. We additionally report hard-pass rate, per-category clause completion, and evidence coverage, with optional breakdowns over realistic, feature-focused, and adversarial task groups.

\paragraph{\taskicon{figure/icons/icon_3Dvideo.png}3D Video Generation.}
\label{app:metric:video}
The evaluator scores five required shot types---\emph{detail}, \emph{hero}, \emph{wide}, \emph{functional}, and \emph{creative}. Let $q_{ij}^{\mathrm{video}}\in[0,1]$ be the quality score for shot type $j\in\{1,\ldots,5\}$. An episode passes only if every required shot scores strictly above $0.60$:
\begin{equation}
    b_i^{\mathrm{video}}
    =\prod_{j=1}^{5}\mathbf{1}\!\left[q_{ij}^{\mathrm{video}}>0.60\right].
\end{equation}

We check the five per-shot quality scores for detail, hero, wide, functional, and creative views. Their visual sub-dimensions include subject visibility, target scale, composition, detail, context, viewpoint diversity, and exposure. Robot-side execution validity is evaluated separately from image quality using completion of all required captures and transitions, inverse-kinematics success, pose error, collision, and minimum clearance.

\paragraph{\taskicon{figure/icons/icon_lego1.png}LEGO Building.}
\label{app:metric:lego-building}
This task requires precisely placing and connecting LEGO elements to realize a specified structure. The evaluator first finds the optimal correspondence between assembled and target bricks, allowing exchanges of identical bricks, global planar translation, and rotations by multiples of $90^\circ$. The episode-level success indicator is
\begin{equation}
    b_i^{\mathrm{LEGO}}
    =G_i^{\mathrm{LEGO}}C_i^{\mathrm{LEGO}}R_i^{\mathrm{LEGO}},
\end{equation}
where $G_i^{\mathrm{LEGO}}=1$ requires every brick to have at most $3\,\mathrm{mm}$ position error and $10^\circ$ orientation error; $C_i^{\mathrm{LEGO}}=1$ requires every expected stud--socket interface to have the correct number of connected studs; and $R_i^{\mathrm{LEGO}}=1$ requires the released assembly to lie on the platform and remain stable without support.

Diagnostic geometry and connection metrics include maximum position and orientation error, stud--socket lateral and depth error, and connector-normal error. Stability duration and post-release drift localize near-miss failures in the released assembly.

\paragraph{\taskicon{figure/icons/icon_lab.png}Automated Biology Laboratory.}
\label{app:metric:biology}
The objective is to complete a requested liquid-handling and measurement procedure while maintaining physical correctness, grounded state tracking, safe operation, and recoverability. Each episode receives the primary score
\begin{equation}
    s_i^{\mathrm{bio}}
    =w_gG_i^{\mathrm{bio}}+w_cC_i^{\mathrm{bio}}
     +w_rR_i^{\mathrm{bio}}+w_eE_i^{\mathrm{bio}}
     +w_sS_i^{\mathrm{bio}}+w_dD_i^{\mathrm{bio}},
\end{equation}
where the six normalized dimensions measure \emph{liquid-state goal completion}, \emph{camera-grounded state verification}, \emph{fault recovery}, \emph{action and resource efficiency}, \emph{safety compliance}, and \emph{low-level command lowering}, respectively. Their weights are $(w_g,w_c,w_r,w_e,w_s,w_d)=(0.30,0.20,0.15,0.15,0.10,0.10)$.

The strict-pass verdict is the conjunction of all hard gates: all scientific goals must be satisfied within tolerance; the terminal tool state must be clean; the episode must be explicitly finished; required camera evidence and device sequences must be complete; no safety violation or unresolved unknown effect may remain; and any required drift recovery must be completed. We retain the six component scores---liquid-state goal completion, camera-grounded state verification, fault recovery, action and resource efficiency, safety compliance, and low-level command lowering---together with hard-gate failure categories as diagnostics.

\paragraph{\taskicon{figure/icons/icon_lego2.png}Functional LEGO Design.}
\label{app:metric:functional-lego}
The objective is to design and build a structure that satisfies an open-ended physical function. After an episode terminates, its final physical state is copied into an independent test rollout in which robot support is removed and task-specific factors such as loads, free objects, impacts, gravity perturbations, or pressing inputs are applied. Let $p_i^k\in\{0,1\}$ denote the outcome of required physical predicate $k$. The primary success indicator is
\begin{equation}
    b_i^{\mathrm{func}}
    =\prod_{k\in\mathcal{P}_i}p_i^k,
\end{equation}
where $\mathcal{P}_i$ is the set of required predicates and $1\leq|\mathcal{P}_i|\leq4$. Thus, all required predicates must hold; partial functional behavior receives no success credit.

Predicate-level outcomes are retained to identify which physical requirement failed. These outcomes cover load support and retention, clearance or passage, protection and interception, routing, containment, separation, and mechanism response. Procedure-based textual completion claims, construction traces, action counts, interface behavior, and observation records are diagnostic only and do not determine functional success.

\paragraph{\taskicon{figure/icons/icon_game.png}Tabletop Games.}
\label{app:metric:games}
Hanabi, Regicide, and Pandemic are evaluated using their native terminal game objectives. Let $g_i^{(m)}$ denote the final game score for episode $i$ in game $m$. The normalized primary score is
\begin{equation}
    s_i^{(m)}=\frac{g_i^{(m)}}{g_{\max}^{(m)}},
    \qquad
    g_{\max}^{(\mathrm{Hanabi})}=25,\quad
    g_{\max}^{(\mathrm{Regicide})}=12,\quad
    g_{\max}^{(\mathrm{Pandemic})}=4.
\end{equation}
For Hanabi, $g_i$ is the sum of the five firework-pile heights; for Regicide, it is the number of enemies defeated; and for Pandemic, it is the number of diseases cured. We report the three game scores separately.

For each game, strict completion means reaching its maximum native score (equivalently, winning Regicide or Pandemic). Decision count, legal-action rate, invalid retries, fallback actions, and physical-execution failures are reported as diagnostic process metrics.

\end{document}